\documentclass[runningheads, envcountsame, a4paper]{llncs}
\usepackage[misc]{ifsym}
\usepackage{cite}
\usepackage{amsmath,amssymb,amsfonts}
\usepackage{graphicx}
\usepackage{epstopdf}
\usepackage{textcomp}
\usepackage{xcolor}
\usepackage{amsmath,tikz}
\usetikzlibrary{matrix}
\usepackage{algorithmic}
\usepackage[ruled, vlined, linesnumbered]{algorithm2e}
\usepackage{caption}
\usepackage{subfigure}
\usepackage{algorithmic}
\usepackage{mathtools}
\usepackage{multirow}
\usepackage{booktabs}
\usepackage[hyphens]{url}

\begin{document}
\title{Spatio-Temporal Tensor Sketching via \\Adaptive Sampling}

\newcommand{\frameworkName}{SkeTenSmooth}
\newcommand{\methodNameSke}{SkeTen}
\newcommand{\methodNameCom}{SkeSmooth}
\newcommand{\partitle}[1]{\smallskip \noindent \textbf{#1.}}
\newcount\Comments   
\Comments=1               
\definecolor{purple}{rgb}{1,0,1}
\newcommand{\kibitz}[2]{\ifnum\Comments=1\textcolor{#1}{#2}\fi}

\author{Jing Ma \Letter \and
Qiuchen Zhang \and
Joyce C. Ho \and Li Xiong}
\authorrunning{J. Ma et al.}
\toctitle{Spatio-Temporal Tensor Sketching via Adaptive Sampling}
\tocauthor{Jing~Ma}
\institute{Department of Computer Science\\
Emory University \\
\email{$\{$jing.ma, qzhan84, joyce.c.ho, lxiong$\}$@emory.edu}}
\maketitle              
\begin{abstract}
Mining massive spatio-temporal data can help a variety of real-world applications such as city capacity planning, event management, and social network analysis. The tensor representation can be used to capture the correlation between space and time and simultaneously exploit the latent structure of the spatial and temporal patterns in an unsupervised fashion. However, the increasing volume of spatio-temporal data has made it prohibitively expensive to store and analyze using tensor factorization. 

In this paper, we propose \frameworkName, a novel tensor factorization framework that uses adaptive sampling to compress the tensor in a temporally streaming fashion and preserves the underlying global structure. \frameworkName~adaptively samples incoming tensor slices according to the detected data dynamics. Thus, the sketches are more representative and informative of the tensor dynamic patterns. In addition, we propose a robust tensor factorization method that can deal with the sketched tensor and recover the original patterns. Experiments on the New York City Yellow Taxi data show that \frameworkName~greatly reduces the memory cost and outperforms random sampling and fixed rate sampling method in terms of retaining the underlying patterns.

\keywords{spatio-temporal data \and tensor sketching \and tensor completion.}
\end{abstract}

\section{Introduction}
The increasing availability of spatio-temporal data has brought new opportunities in application domains including urban planning, event management, and informed driving  \cite{gauvin2014detecting,sun2016understanding}.
Unfortunately, the rapid growth in these data streams can be prohibitively expensive to store, communicate and analyze.
In addition, the high-dimensional, multi-aspect spatio-temporal data poses analytic challenges due to the correlations in the measurements from both time and space.
Moreover, human-intensive and domain-specific supervised models are not tractable due to the constant and evolving deluge of measurements.

Given the high-dimensional, multi-aspect nature of spatio-temporal data, a tensor serves as an efficient way to represent and model such data \cite{sun2016understanding,gauvin2014detecting,chen2016fine}. 
As an example, each element of the tensor can represent the occurrences of an event at a specific location (encoded as latitude and longitude) within a specific time interval.
Compared with the low-dimensional matrix-based methods such as \cite{lakhina2004structural}, tensors can capture the correlation between each mode. Furthermore, tensor factorization offers an unsupervised, data-driven approach to identify the global structure of the data via a high-order decomposition. It is also more interpretable compared to deep learning methods \cite{yokota2016smooth}.
Unfortunately, existing models require the full tensor information (i.e., all the data samples must be stored) and do not readily scale to extremely large tensors.

The computation and storage limitations of large tensors motivate the need to approximate such tensors with relatively small ``sketches" of the original tensors.
Not only are these manageable-sized tensor sketches more readily stored on a single machine, the computationally expensive tasks such as tensor decomposition can be performed on the smaller tensors while still preserving the underlying structure.
Although sketching has been proposed as a linear algebra tool for reducing the memory cost, traditional methods involve the linear transformation of the original tensor with a ``fat" random projection matrix to reduce the dimensions \cite{woodruff2014sketching,sidiropoulos2014parallel}.
However, the transformation has inherent shortcomings: 1) the design of the random projection matrix may not capture the evolving patterns without a priori information; and 2) continuously sketching is computationally expensive since it involves the inversion of a fat random projection matrix.

An alternative sketching approach is based on tensor sparsification -- subsampling the original tensor while preserving the original tensor structure.
Compared with the random projection method, sampling incurs negligible online complexity. Existing tensor sparsification methods include random sampling based on the tensor spectral norm \cite{nguyen2010tensor}, sampling according to the entry values \cite{xia2017effective}, and sampling according to the pre-computed tensor distribution \cite{bhojanapalli2015new}. 
However, these tensor sparsification algorithms suffer from the following limitations: 1) they cannot deal with streaming data where measurements are not available a priori and the tensor is incrementally updated; 2) there is no formal mechanism to reconstruct the original streams \cite{gama2016rethinking}.
While tensor reconstruction is often a byproduct of dealing with missing data, reconstruction of the original streams can help track dynamic and abrupt changes at particular locations.

\begin{figure}[h]
\centering{\includegraphics[width=3in]{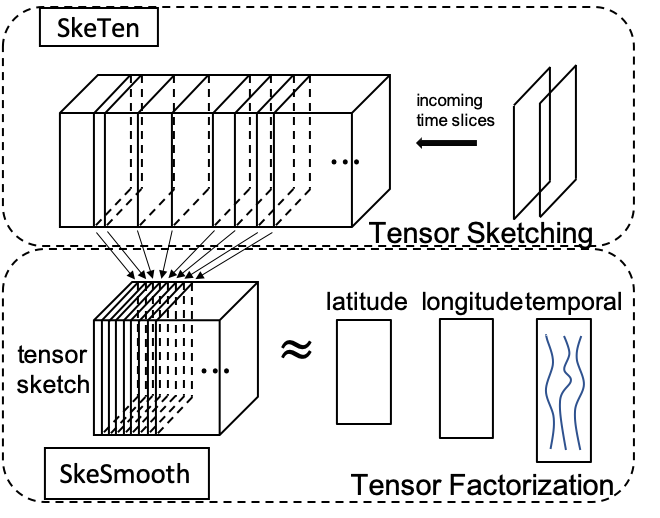}}
\caption{The overall flow of \frameworkName, including \methodNameSke~and \methodNameCom.}
\label{fig:framework}
\end{figure}

In this paper, we propose \frameworkName, a factorization framework that uses adaptive sampling to generate tensor sketches on-the-fly and preserves the underlying global structure.
We explore the problem of serially acquired time slices (measurements appear once they are available). We introduce \methodNameSke, a tensor sketching method that adaptively adjusts the sampling intervals and samples time slices according to the feedback error between the prior estimate and the true value.
Thus it can capture the time slices that are not well modeled by the prior forecasting model, while avoiding the storage of the time slices that contain redundant information (i.e., patterns that are already captured). Furthermore, we propose a novel method \methodNameCom~to decompose the tensor sketches and reconstruct the underlying temporal trends.
Unlike previous tensor factorization algorithms that deal with randomly missing entries, the sketched tensor has missing time slices (i.e., no data from a specific time point) which poses additional challenges.
Thus, we introduce a robust tensor factorization algorithm that incorporates temporal smoothness constraints using auxiliary information about the data gained from \methodNameSke.
Figure \ref{fig:framework} shows the overall flow of the proposed framework.
We briefly summarize our contributions as:

1) \textbf{Tensor sketching with adaptive sampling.} We propose \methodNameSke, a tensor sketching technique that helps process large volumes of data with low memory requirements and preserves the underlying temporal dynamics.

2) \textbf{Tensor factorization with smoothness constraint.} We propose a tensor factorization framework, \methodNameCom, that decomposes the smaller tensor ``sketches" with smoothness constraints to achieve robust recovery of the underlying latent structure and the missing entries. 

3) \textbf{Case study on New York taxi data.} We illustrate the ability to produce small tensor ``sketches" and generate smooth temporal factors on real data.

\section{Preliminaries and notations}
This section summarizes the notations used in this paper. Note that we use $\mathcal{X}$ to denote a tensor, \textbf{X} to denote a matrix, and $\textbf{x}$ to denote a vector. The mode-$d$ matricization of a tensor is denoted by $X_{(n)}$. The row and column vectors are represented by $\textbf{x}_{i:}, \textbf{x}_{:r}$ respectively.

\subsection{Tensor Factorization}

\begin{definition}
(\textit{Khatri-Rao product}). Khatri-Rao product is the ``columnwise" Kronecker product of two matrices $\mathbf{A}\in \mathbb{R}^{I\times R}$ and $\mathbf{B}\in \mathbb{R}^{J\times R}$. The result is a matrix of size $(IJ\times R)$ and defined by
$
\mathbf{A}\odot \mathbf{B}=\left[\mathbf{a}_1\otimes \mathbf{b}_1\cdots \mathbf{a}_R\otimes \mathbf{b}_R\right]
$.
$\otimes$ denotes the \textit{Kronecker product}. The \textit{Kronecker product} of two vectors $\mathbf{a}\in \mathbb{R}^{I}$, $\mathbf{b}\in \mathbb{R}^{J}$ is 
$$\mathbf{a}\otimes \mathbf{b}=\left[\begin{array}{c}
a_1 \mathbf{b}\\
\vdots\\
a_I \mathbf{b}
\end{array}\right]$$
\end{definition} 

\begin{definition}
(\textit{CANDECOMP-PARAFAC Decomposition}). 
The CANDECOMP-PARAFAC (CP) decomposition is to approximate the original tensor $\mathcal{Y}$ by the sum of $R$ rank-one tensors where $R$ is the rank of tensor $\mathcal{Y}$.
For a three-mode tensor $\mathcal{Y}\in \mathbb{R}^{I\times J\times K}$, the CP decomposition can be represented as
\begin{equation}
\mathcal{O} \approx \mathcal{X}=\sum\limits_{r=1}^{R}\textbf{a}_{:r}\circ \textbf{b}_{:r}\circ \textbf{c}_{:r},
\end{equation}
where $\textbf{a}_{:r}\in \mathbb{R}^I$, $\textbf{b}_{:r}\in \mathbb{R}^J$, $\textbf{c}_{:r}\in \mathbb{R}^K$ are the $r$-th column vectors within the three factor matrices $\textbf{A}\in \mathbb{R}^{I\times R}$, $\textbf{B}\in \mathbb{R}^{J\times R}$, $\textbf{C}\in \mathbb{R}^{K\times R}$, $\circ$ denotes the outer product.
\end{definition}
In this paper, the spatio-temporal tensor has three dimensions. The first dimension is the temporal dimension, while the other two are longitude and latitude dimensions representing the spatial modes. 

\section{\methodNameSke: Tensor Sketching via Adaptive Sampling}
The main idea of \methodNameSke~is to discard slices along the temporal mode that are ``predictable".
Intuitively, this adaptive sampling strategy will capture sudden temporal changes of the incoming streams, which is difficult for other methods such as random sampling and fixed rate sampling to capture.
Moreover, \methodNameSke~does not require a priori knowledge of the data and can adjust the sampling interval in real-time.
\methodNameSke~consists of two main components. The adaptive sampling module uses a prediction model to measure the ``predictability" of the data. 
The second component selects random time series projections to reduce the training time of the prediction models.

\subsection{Adaptive Sampling}

\methodNameSke~introduces an adaptive sampling method that can adjust according to the detected temporal dynamics and thus can better capture the underlying patterns even with the same amount of data stored. 
In \methodNameSke, individual time-series prediction models are developed for each spatial location. 
The adaptive sampling strategy then analyzes the slice feedback error, or the feedback error across all the locations.

\begin{definition}
(Slice Feedback Error). Each tensor fiber along the temporal mode, $\mathcal{X}(:,j,k)$, is denoted as a time-series stream, $\textbf{x}_{j,k}$. If $\textbf{x}_{j,k}^{(t)}$ represents the true value at a particular time $t$, where $0\leq t\leq T$, then the slice feedback error at time $t$, $E_t$ is defined as: 
\begin{equation}
    E_t=\sum\limits_{k=1}^{K}\sum\limits_{j=1}^{J}{\left|\hat{\textbf{x}}_{j,k}^{(t)}-\textbf{x}_{j,k}^{(t)}\right|}/\max\{\textbf{x}_{j,k}^{(t)},\delta\},
    \label{eq:feedback}
\end{equation}
where $\hat{\textbf{x}}_{j,k}^{(t)}$ is the estimated value based on the prediction model, and $\delta$ is a small sanity bound.
\end{definition}

The slice feedback error reflects how well the current time series models fit the current trends for each time series fiber.  
If there is a sudden increase across multiple spatial locations (i.e., an increase in the slice feedback error), then the sampling interval should be shortened to better capture the evolving trends.

\subsubsection{Time Series Prediction Model}
\methodNameSke~uses an Autoregressive Integrated Moving Average (ARIMA) model to predict the values of the temporal slices.
Given a time-series sequence $\{x_i\}$, $i=1,\cdots,t$, where $i$ is the time index, ARIMA($p,d,q$) is defined as 
\begin{equation}
    \nabla^d x_t = \sum\limits_{j=1}^{p} \alpha_j x_{t-j} + \varepsilon_t +  \sum\limits_{j=1}^{q}\beta_j \varepsilon_{t-j} + c
         \label{eq:arima}
\end{equation}
where $p$ is the order of lags of the autoregressive model, $d$ is the degree of differencing, and $q$ is the order of the moving average model. $\alpha_j, j=1,\cdots,p$ and $\beta_j, j=1,\cdots,q$ represent the coefficients of AR and MA, respectively, $\varepsilon_t$ is the white noise at time $t$, and $c$ is the bias term.
An ARIMA model is trained on each tensor fiber 
for the initial sampled time slices.

ARIMA has several convenient properties that make it more suitable than other time series prediction models in our setting. (1) The auto-regressive part of ARIMA ($\{\alpha_1,\cdots,\alpha_p\}$) generates the coefficients that are later exploited in \methodNameCom~for the smoothness constraint in Sec. \ref{sec:skesmooth}.  This is essential for recovering the original temporal patterns with missing time slices. (2) It doesn’t require a large number of training samples so the compression can occur much earlier than more complex models. (3) The coefficients can be used to understand the adaptive sampling for the PID controller which we will explain next.
In preliminary experiments, we tried several state-of-the-art deep learning models such as ConvLSTM \cite{xingjian2015convolutional}, but the compression rates and sampled slices only improved marginally.
Thus, ARIMA is an optimal choice for our setting.

\subsubsection{Feedback Control}

\methodNameSke~uses a controller system to detect rapid changes in the slice feedback error and adaptively adjusts the sampling rate.
We adopt a PID controller similar to \cite{fan2013adaptive,king2011process} to change the sampling interval over time. The PID controls the \textit{Proportional, Integral} and \textit{Derivative} errors. \textit{Proportional error} is defined as $\Gamma_p=\gamma_p E_t$ to control the sampling intervals by keeping the controller proportional to the current slice feedback error. \textit{Integral error} is defined as $\Gamma_i={\frac{\gamma_i}{M_t}}\sum_{m=0}^{M_t}E_t$, where $M_t$ represents how many errors have been taken until time $t$. Thus the integral error considers the past errors to eliminate offset. \textit{Derivative error} is defined as $\Gamma_d=\gamma_d{\frac{E_t-E_{t-1}}{t_m-t_{m-1}}}$ to prevent large errors in the future. Therefore, the full PID controller is defined as
\begin{align}
    ~&\Gamma=\Gamma_p+\Gamma_i+\Gamma_d, \notag\\
    \text{s.t.}~& \gamma_p,\gamma_i,\gamma_d\geq0, \ \gamma_p+\gamma_i+\gamma_d=1.
    \label{eq:pid}
\end{align}

The PID errors can thus be interpreted as control actions based on the past, the present and the future \cite{aastrom2013computer}. We set the PID parameters according to the common practice that $proportional > integral >
derivative$ \cite{fan2013adaptive}. With the PID error, the sampling interval is adjusted according to
\begin{equation}
    newInterval=\max\{1,\theta(1-e^{\frac{\Gamma-\xi}{\xi}})\},
    \label{eq:interval}
\end{equation}
where $\theta$ and $\xi$ are predefined user-specific parameters. The adaptive sampling process is presented in Algorithm \ref{algo1}. 

\begin{algorithm}[t]
\DontPrintSemicolon
\KwIn{ARIMA models for each tensor fiber, incoming tensor slices $\mathcal{X}(:,j,k),(j\leq J, k\leq K)$, set of PID controller parameters $\{\gamma_p,\gamma_i,\gamma_d\}$, next sampling point $ns$}
\While{not reach the end of data streams}
{
\If{next time slice $t==ns$}{
    sample the next time slice to the sketched tensor;\\
    compute slice feedback error according to eq. (\ref{eq:feedback});\\
    compute the PID controller error $\Gamma$ according to eq. (\ref{eq:pid});\\
    obtain the new interval according to eq. (\ref{eq:interval});\\
    update the next sampling point $ns=ns + newInterval$;\\
    }
    \Else{
    move on to the next time slice.
    }
}\label{endwhile}
\caption{\methodNameSke}\label{algo1}
\end{algorithm}

\subsection{Random Projection of ARIMA Coefficients}
For large scale spatio-temporal data with many spatial locations (countless values of longitude and latitude), training the prediction model (i.e., ARIMA) may be computationally infeasible. To reduce the computational cost and retain the temporal fidelity of the data, we propose a random projection algorithm to train a limited number of ARIMA models, while preserving the competitive performance of the trained models. 
Inspired by the Locality Sensitive Hashing (LSH) algorithm, which has demonstrated its success in Approximate Nearest Neighbor search, our proposed projection method has the following steps: 
(1) train $L$ ARIMA models $g_i, i=1,...,L$ from $L$ time series $ts_i, i=1,...,L$, where each of the $L$ time series are chosen at random from the time series set $\mathcal{T}$ containing all tensor fibers $ts_j, j=1,...,M$; (2) construct $L$ buckets from the time series set $\mathcal{T}$, each bucket $i$ corresponds to an ARIMA model $g_i$; (3) map each time series $ts_j, j=1,...,M$ from $\mathcal{T}$ into bucket $i$ according to the RMSE between the true and the predicted values using the ARIMA model $g_i$. In this way, we treat the fitting of each time series to the randomly selected time series as a special ``random projection", and then map each time series according to the least RMSE to each bucket $i$. This process (detailed in Algorithm \ref{algo:randproj}) 
allows locations that may not have been spatially close to be clustered to the same model and can decrease the training time, which is the bottleneck of the adaptive sampling process. 

We compare our proposed random projection algorithm for the ARIMA coefficients with the k-means algorithm where similar time series can also be clustered and thereby reducing training time. Fig. \ref{fig:theta} (c), (d) demonstrate that despite the high computation cost of the k-means algorithm, it provided limited improvement over the random projection algorithm. 

\begin{algorithm}[t]
\DontPrintSemicolon
\KwIn{Tensor fiber set $\mathcal{T}=\{ts_1,...,ts_M\}$, number of models to train $L$}
\For{i=1,...,L}
{
  train ARIMA model $g_i$ for the randomly selected time series $ts_i$
}
  \For{j=1,...,M}
  {
    \For{i=1,...,L}
    {
      map each $ts_j$ into bucket $i$ according to $arg\min\limits_{g_i}||g_i(ts_j)-ts_j||$
    }
  }\label{endfor}
\caption{Random Projection of ARIMA Coefficients}\label{algo:randproj}
\end{algorithm}

\section{\methodNameCom: Smooth Tensor Factorization}

We then consider analyzing the sketched tensor using the CP decomposition to capture the underlying multi-linear latent structure of the data and reconstruct the unsampled time slices. 
Built on the success of the Quadratic Variation (QV) matrix constraint \cite{yokota2016smooth,zhou2015spatio}, we consider the delay effect of the time series and utilize the coefficients from the pre-trained ARIMA model from \methodNameSke~to make the temporal pattern smooth and robust to missing entire time slices.

\subsection{Formulation}
\label{sec:skesmooth}
\partitle{CP Decomposition with Missing Data}
We consider the sketched tensor as an incomplete data problem where the goal is to both learn the underlying patterns of the data and reconstruct the unsampled values.
Previous CP-based tensor completion algorithms \cite{acar2011scalable,song2019tensor} have formulated the completed tensor $\mathcal{\overline X}$ as:
\begin{equation}
    \mathcal{\overline X} = \mathcal{W}*\mathcal{X}+(1-\mathcal{W})*{[\![\textbf{A},\textbf{B},\textbf{C}]\!]},
\end{equation} where $*$ denotes the element-wise product, the observed tensor is factorized as $\mathcal{X}=[\![\textbf{A},\textbf{B},\textbf{C}]\!]$ and the binary weight tensor $\mathcal{W}$ is of the same size as $\mathcal{X}$ such that
\begin{equation*}
    w_{i,j,k} = 
    \left\{
        \begin{array}{lr}
            1, \quad \text{if $w_{i,j,k}$ is known} &  \\
            0, \quad \text{if $w_{i,j,k}$ is missing}. &  
        \end{array}
\right.
\end{equation*}

\partitle{ARIMA-based Regularization Matrix}
Standard CP decomposition with missing data assumes the entries are missing at random, whereas \methodNameSke~yields a sketched tensor with entire time slices missing.
Moreover, they do not account for temporal information  which can improve interpretability and robustness.
In spatio-temporal datasets, successive observations at the same spatial location (i.e., $x_{j,k}^{(t)}$ and $x_{j,k}^{(t+1)}$) are unlikely to change significantly.
Thus, adjacent time slots are generally smooth.
\methodNameCom~incorporates the coefficients generated by the various ARIMA models as auxiliary information into the regularization matrix $\mathcal{L}$.
Based on Eq. \eqref{eq:arima}, we observe that $\textbf{a}_i$ can be approximated as: 
\begin{equation}
\textbf{a}_i=\alpha_1\textbf{a}_{i-1}+\alpha_2\textbf{a}_{i-2}+\cdots+\alpha_p\textbf{a}_{i-p},~p<i
\end{equation}
Thus by utilizing the learned predictive models in \methodNameSke~(the autoregressive coefficients and the order of the number of time lags, $p$), the algorithm can better reconstruct the missing time slices and learn the underlying patterns.

Although each ARIMA model may have different order of time lags, $p_{j,k}$, we use the largest order $p = \max p_{j,k}$ and set the coefficients $\alpha_{p_{j,k}+1}, \cdots, \alpha_p,$ to zero if $p_{j,k} < p$.
Since each tensor fiber is clustered to the ARIMA model that fits best, we can thus get the weight for each ARIMA model as the number of tensor fibers clustered to it. We then compute the weighted average of the autoregressive coefficients to produce the regularization matrix, $\mathcal{L}$, defined as:
\begin{equation}
    \mathcal{L}=
    \begin{bmatrix}
        -\alpha_p  & 0 & 0 &  0  & \cdots & 0  \\
        \vdots & \ddots & \vdots & \vdots & \ddots & \vdots \\
        -\alpha_1  & \cdots & -\alpha_p & 0 & \cdots & 0 \\
        1 & -\alpha_1 & \cdots & -\alpha_p & \cdots & 0 \\
        \vdots & \ddots & \ddots & \vdots & \ddots & \vdots\\
        0  & \cdots & 1 & -\alpha_1  & \cdots & -\alpha_p \\
    \end{bmatrix}
\end{equation}

\partitle{\methodNameCom~Optimization Problem}
The objective function for \methodNameCom~after adding the smoothness constraint is:
\begin{align}
\label{eq:TFSmooth}
    \min \limits_{\mathcal{X}} \frac{1}{2} {\left\| \mathcal{Y} - \mathcal{W}*{[\![\textbf{A},\textbf{B},\textbf{C}]\!]}\right\|}_F^2 + \frac{\rho}{2} {\left\|\mathcal{L}\textbf{A}\right\|}_F^2, \quad   
    \text{s.t.~} \mathcal{Y}=\mathcal{W}*\mathcal{X},
\end{align}
where $\mathcal{Y}$ denotes the sketched tensor with missing entries.
Thus higher values of the regularization parameter, $\rho$, will yield smoother temporal factors. However, this may potentially reduce the overall fit of the tensor.

\subsection{Algorithm}

We use a first-order method to solve the optimization problem, similar to \cite{acar2011scalable}. The gradient-based method has been shown to be robust to overspecification of the rank.
As the regularization matrix is only enforced on the temporal mode (first mode), only the gradient for the temporal mode involves the gradient of the regularization matrix.
Thus the updates for all three modes are:
\begin{align}
    ~& \textbf{G}^{(1)}=-(\mathcal{Y}_{(1)}-\mathcal{Z}_{(1)})(\textbf{C}\odot\textbf{B}) + \rho \mathcal{L}^{T}\mathcal{L}\textbf{A}; \notag \\
    ~& \textbf{G}^{(2)}=-(\mathcal{Y}_{(2)}-\mathcal{Z}_{(2)})(\textbf{C}\odot\textbf{A}); \notag\\
    ~& \textbf{G}^{(3)}=-(\mathcal{Y}_{(3)}-\mathcal{Z}_{(3)})(\textbf{B}\odot\textbf{A});\notag\\
    \text{s.t.}~& \mathcal{Z} = \mathcal{W}*{[\![\textbf{A},\textbf{B},\textbf{C}]\!]},
    \label{eq:grad}
\end{align}
where $\mathcal{Y}_{(n)}$ is the matricization on the $n$-th mode of the tensor sketch.
Our gradient-based algorithm uses the limited-memory BFGS \cite{nocedal2006numerical} for the first order optimization. 
The optimization process is shown in Algorithm \ref{algo3}.

\subsection{Complexity Analysis}
The time complexity of \methodNameCom~is dominated by the matricized tensor times Khatri-Rao product (MTTKRP) operations, which is computed in Eq. (\ref{eq:grad}) as the matricized residual tensor $\mathcal{Y}_{(n)}-\mathcal{Z}_{(n)}$ times the Khatri-Rao product between two factor matrices. For simplicity purpose, we denote an N-th order tensor $\mathcal{X}\in \mathbb{R}^{I\times\cdots \times I}$ as the representation of the residual tensor $\mathcal{Y}_{(n)}-\mathcal{Z}_{(n)}$. For each mode, computing MTTKRP takes $O(nnz(\mathcal{X})R)$, where $R$ is the approximate rank of the tensor, $nnz(\mathcal{X})$ represents the number of non-zero elements of $\mathcal{X}$. While computing the term $\rho \mathcal{L}^{T}\mathcal{L}\textbf{A}$ for the first mode takes $O(I^2R^2)$. Thus the overall time complexity for \methodNameCom~is  $O(nnz(\mathcal{X})NR+I^2R^2)$.

\begin{algorithm}[t]
\DontPrintSemicolon
\KwIn{Sparse tensor sketch $\mathcal{Y}$, weight tensor $\mathcal{W}\in \mathbb{R}^{I\times J\times K}$, smooth parameter $\rho$.}
Randomly initialize the factor matrices $\textbf{A}\in \mathbb{R}^{I\times R},\textbf{B}\in \mathbb{R}^{J\times R},\textbf{C}\in \mathbb{R}^{K\times R}$\\
\While{$\textbf{A},\textbf{B},\textbf{C}$ not converge}
{
    compute tensor $\mathcal{Z}=\mathcal{W}*{[\![\textbf{A},\textbf{B},\textbf{C}]\!]}$;\\
    compute function value $f = \frac{1}{2} {\left\| \mathcal{Y} - \mathcal{Z}\right\|}_F^2$;\\
    update gradient $G^{n}$ according to Eq. \eqref{eq:grad};\\
    update $\textbf{A}$, $\textbf{B}$ and $\textbf{C}$ with l-bfgs \cite{nocedal2006numerical}.\\
}\label{endwhile}
\caption{\methodNameCom}\label{algo3}
\end{algorithm}

\section{Experiments}
We evaluate \frameworkName~on a large real-world dataset, the New York City (NYC) Yellow Taxi data (see Supplementary for experiment settings). The goal of our evaluation is to assess both \methodNameSke~and \methodNameCom~from three aspects:

\begin{enumerate}
    \item \textbf{Effectiveness of \methodNameSke:} Analyze memory compression without hurting the performance. Use \methodNameSke~to evaluate the latent structure preservation.
    \item \textbf{Effectiveness of \methodNameCom:} Evaluate with the sketched tensor on the effectiveness of the smoothness constraints.
    \item \textbf{Sketching result on prediction tasks:} Analyze how the decomposed factor matrices perform on a downstream prediction task, to indirectly evaluate whether the sketching results are meaningful.
\end{enumerate}

\subsection{Dataset}
We evaluate \frameworkName~on the NYC Yellow Taxi Data\footnote{https://www1.nyc.gov/site/tlc/about/tlc-trip-record-data.page}, which is collected and published by the New York City Taxi and Limousine Commission. To demonstrate the scalability of our methods, we collect data from January, 2009 to June, 2015, which has 1,090,939,222 trips in total, on a daily basis\footnote{We also conducted experiments on hourly basis (a finer granularity), which shows consistent results with the daily basis. Results can be found in Supplementary.}. We investigate the NYC area using the latitude range of $[40.66,40.86]$ and longitude range of $[-74.03,-73.91]$. We partition the NYC area with a $0.001\times 0.001$ degree grid, where 0.001 degree is roughly 111.32 meters. Then we form a tensor $\mathcal{X}$ of size $2341\times120\times200$, for time, latitude and longitude modes respectively. The first 60\% of the tensor is used for training and is thus considered as offline data, while the latter 40\% is used to test \methodNameSke~and \methodNameCom.

\subsection{Evaluation Metrics}
The tensor decomposition performance is evaluated by two criteria: Factor Match Scores (FMS) and Tensor Completion Scores (TCS) \cite{acar2011scalable}. FMS is defined as: 
$$
score(\Bar{\mathcal{X}}) = {\frac{1}{R}}\sum\limits_r {\left( 1-{\frac{\left| \xi_r- \Bar{\xi_r} \right|}{max\left\{\xi_r, \Bar{\xi_r} \right\}}} \right)}\prod\limits_{\textbf{x}=\textbf{a,b,c}}{\frac{\textbf{x}_r^T \Bar{\textbf{x}}_r}{{\left\| \textbf{x}_r \right\|}{\left\| \Bar{\textbf{x}}_r \right\|}}},
$$
$$
\xi_r = \prod\limits_{\textbf{x}=\textbf{a,b,c}}{\left\| \textbf{x}_r \right\|}, \Bar{\xi_r} = \prod\limits_{\textbf{x}=\textbf{a,b,c}}{\left\| \Bar{\textbf{x}}_r \right\|}
$$
where $\Bar{\mathcal{X}} = [\![\Bar{\textbf{A}},\Bar{\textbf{B}},\Bar{\textbf{C}}]\!]$ is the estimated factor and $\mathcal{X} = [\![\textbf{A},\textbf{B},\textbf{C}]\!]$ is the true factor. $\textbf{x}_r$ is the $r^{th}$ column of factor matrices. FMS measures the similarity between two tensor decomposition solutions. It ranges from 0 to 1, and the best possible FMS is 1. We choose the factorization result of the complete tensor using CP-OPT \cite{acar2011scalable} as a true factor to compare with.

TCS indicates the relative error of the missing entries, and is defined as
$$
TCS = {\frac{\left\|(1-\mathcal{W})*(\mathcal{X}-\bar{\mathcal{X}})\right\|}{\left\|(1-\mathcal{W})*\mathcal{X}\right\|}},
$$
The best possible TCS is 0.

\subsection{Parameter Selection}
\begin{figure*}[t]
\centering
\subfigure[Sampling rate vs $\theta$]{
\includegraphics[width=1.05in]{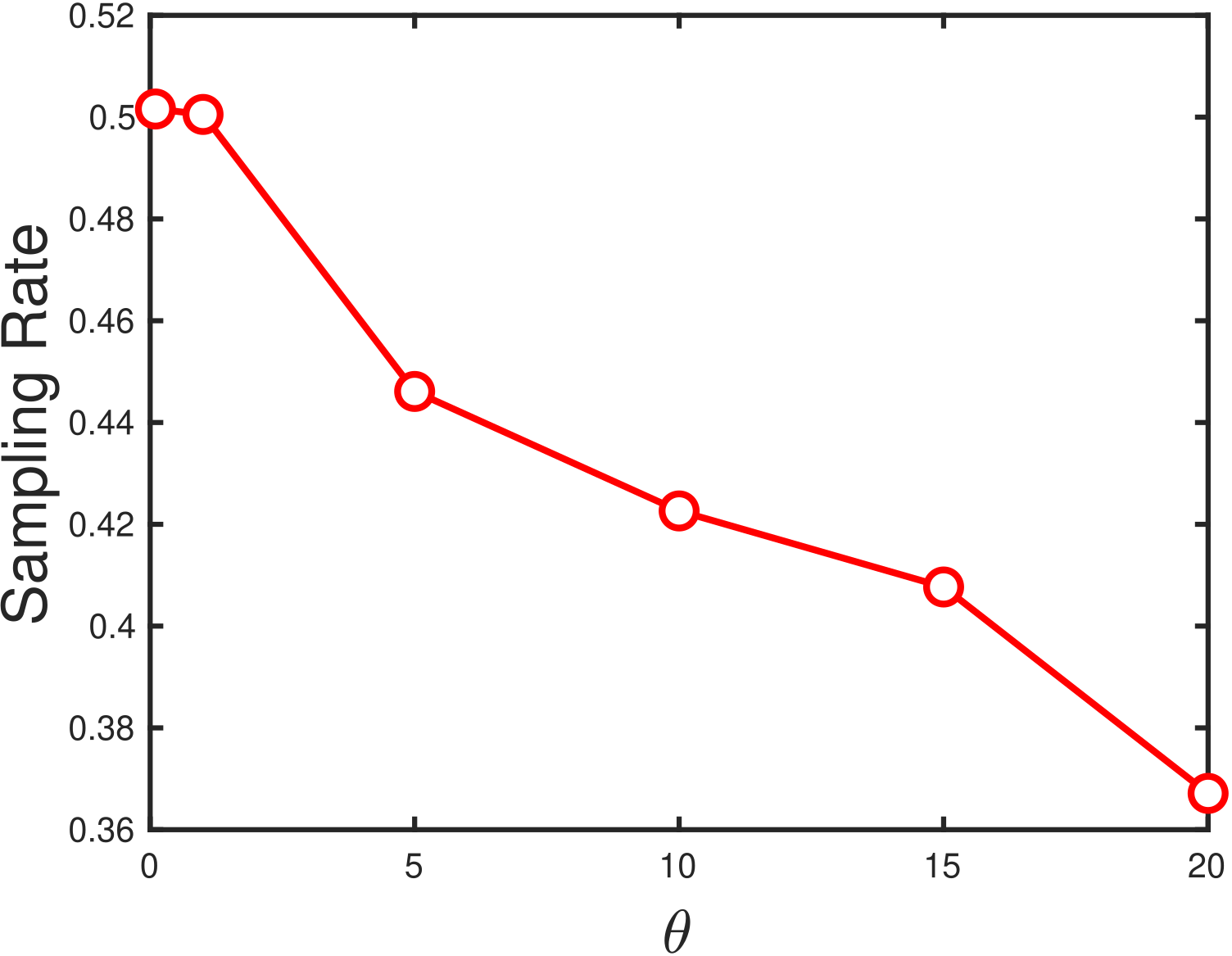}
}
\subfigure[FMS/TCS vs $\theta$]{
\includegraphics[width=1.05in]{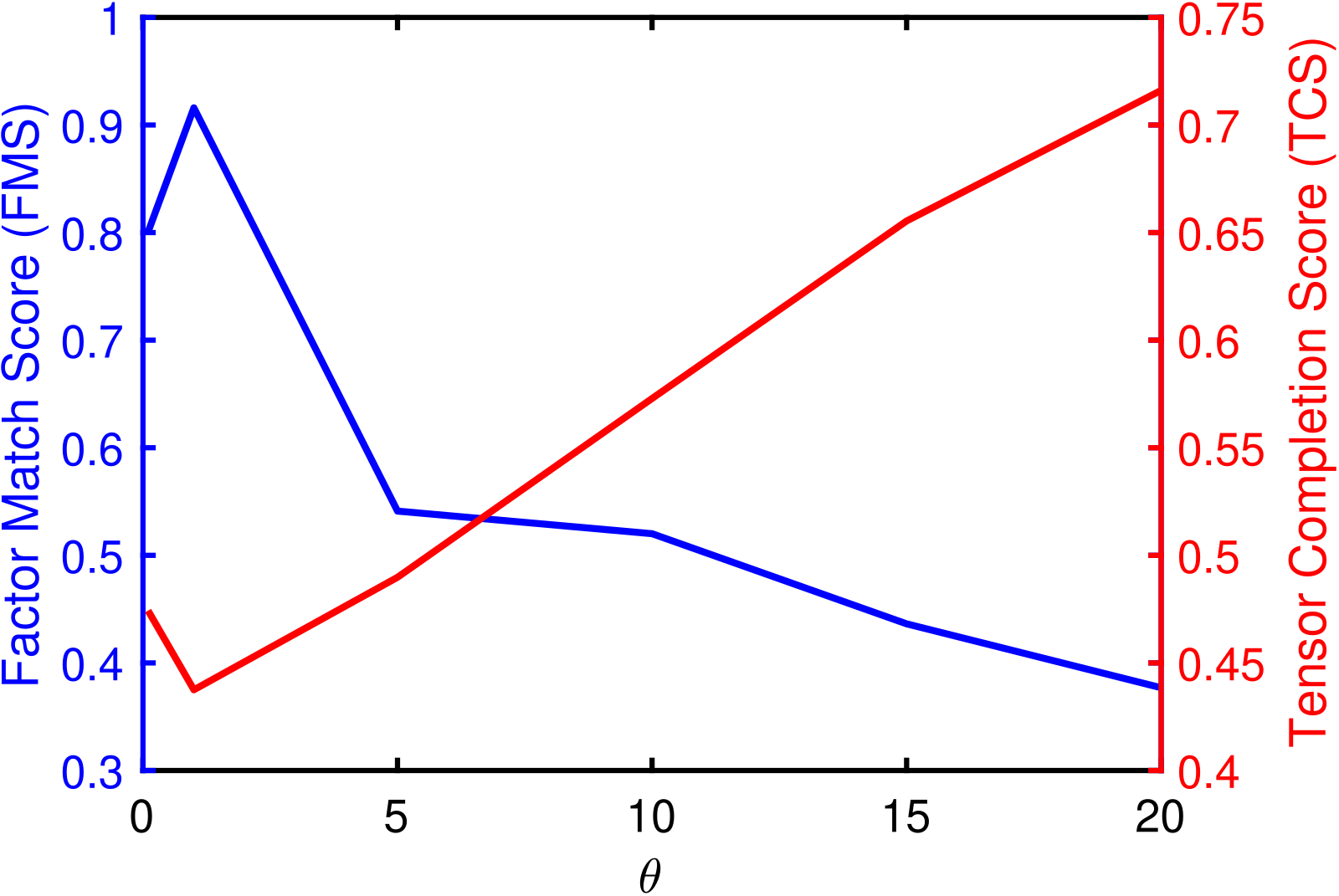}
}
\subfigure[FMS vs dropped rates]{
\includegraphics[width=1.05in]{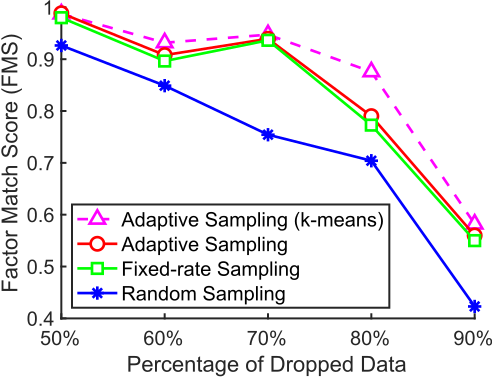}
}
\subfigure[TCS vs dropped rates]{
\includegraphics[width=1.05in]{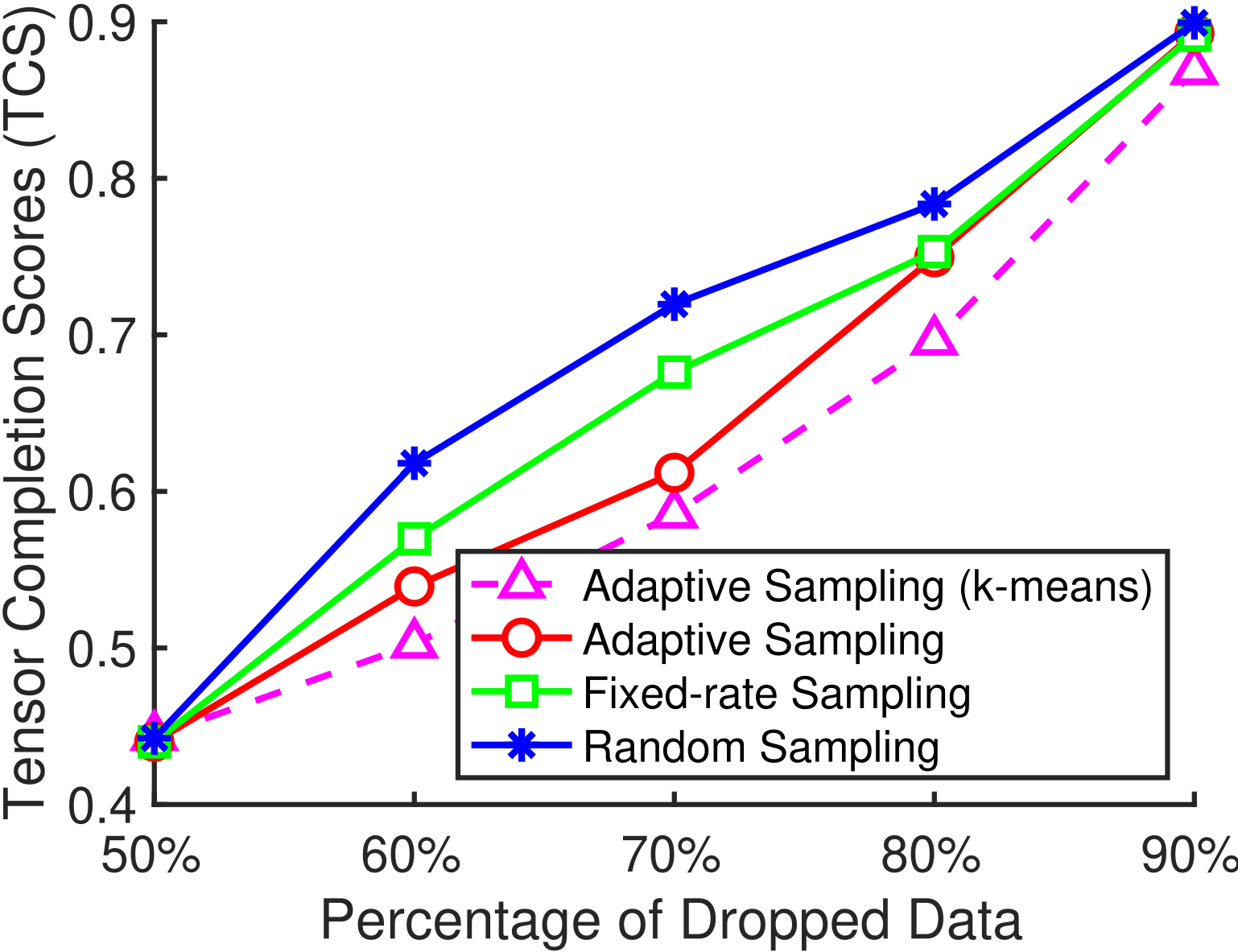}
}
\caption{Overall sampling rates, FMS and TCS over the change of $\theta$ ((a) and (b)). FMS and TCS over dropped rates ((c) and (d)).}
\label{fig:theta}
\end{figure*}

In \methodNameSke, parameters involved in achieving the adaptive sampling algorithm include: the PID controller parameters $\{\gamma_p,\gamma_i,\gamma_d\}$, $\theta$ and $\xi$ for the interval adjustment. The PID parameters are set as $\{\gamma_p,\gamma_i,\gamma_d\}=\{0.7,0.2,0.1\}$\footnote{We experimented with several PID settings, and it hardly affected the sampling rate as long as we set the controller ID according to $proportional > integral >
derivative$. We thus consider our system robust to the control parameters and choose the optimal setting in empirical studies.}. 
Parameter $\theta$ determines the magnitude of the sampling interval adaptation, its impact is shown in fig. \ref{fig:theta}(a) and (b). From fig. \ref{fig:theta}(b), we observe that as $\theta$ increases, the reconstruction performance measured by FMS and the TCS decreases due to the enlarged average sampling intervals (shown in fig. \ref{fig:theta}(a)). The optimal choice shown in the figure is $\theta=1$. When $\theta=0.1$, the reconstruction performance (TCS and FMS) is worse than when $\theta=1$ due to an insufficient interval adjustment. $\xi$ is determined as 3.5 for the tolerance of PID error. 

For \methodNameCom, the smoothness constraint $\rho$ is chosen as 600 after grid search through \{500, 600, 700, 800, 900, 1000\}, and the rank of the tensor is determined as 5 after grid search in \{5, 10, 15, 20, 25\} (see Supplementary for more details).
The number of lags is determined as $p=3$ based on the majority of all the trained ARIMA models. We use the number of tensor fibers been mapped into the same bucket as weights, and compute the final coefficients for the regularization by weighted average of the  coefficients of the randomly trained ARIMA models and is determined as $\boldsymbol{\alpha}= [0.55, -0.19, 0.04]$\footnote{The k-means based coefficients are computed the same way using the Dynamic Time Warping (DTW) distance and is determined as $\boldsymbol{\alpha_{k-means}}= [0.51, -0.07, 0.007]$.}.

\subsection{Tensor Sketching}
Here we consider a streaming setup where the tensor sketching will be applied to a succession of incoming time slices. 
\subsubsection{Baselines}
We compare \methodNameSke~with two baseline sampling techniques:
\begin{itemize}
\item \textbf{Fixed rate sampling} samples the data with a predefined, fixed interval.
\item \textbf{Random sampling} has been proposed as a way to sparsify the tensor \cite{nguyen2010tensor}.
\end{itemize}
By evaluating the performance of fixed rate sampling and random sampling, we can gain a better understanding of the benefits of adaptive sampling under the setting of time-evolving streams and how it may provide more informative representations of the streams at the same storage cost. 

\subsubsection{Evaluation}
\label{sec:eval}
We evaluate the three sampling methods for different levels of dropped data by using the proposed \methodNameCom~algorithm. The performance is measured by both FMS and TCS. We explore the sketching level from 50\% to 90\%. Fig. \ref{fig:theta} (c) and (d) show the FMS and TCS of different sampling methods when applied to \methodNameCom~algorithm. As more data are sampled, FMS gradually approaches 1, which means the extracted temporal patterns are equivalent to the true patterns. Notice that when the level of sampling is 50\%, adaptive sampling and fixed-rate sampling method show FMS as high as 0.9886 and 0.9803, which means that with only 50\% of the original data, \methodNameCom~can successfully recover the original temporal pattern. Adaptive sampling achieves slightly better reconstruction performance than fixed-rate sampling, because the daily taxi data are fairly regular and periodical so the interval adjustments for adaptive sampling helps marginally. Both adaptive sampling and fixed-rate sampling are better than the random sampling, illustrating that carefully designing the sampling strategy is critical for pattern and data reconstruction.

\subsection{Tensor Decomposition with Sketches}
To evaluate the performance of tensor sketching, we first need to look into how it can preserve the temporal trend and the latent structures on the temporal dimension. We decompose the sketched results to generate the factor matrices as latent patterns, and use these factor matrices to reconstruct the original tensor. We compare \methodNameCom~with the following state-of-the-art methods: CP-WOPT \cite{acar2011scalable}, HaLRTC \cite{liu2012tensor}, SPC \cite{yokota2016smooth}, BCPF \cite{zhao2015bayesian}, t-SVD \cite{zhang2014novel}, t-TNN \cite{hu2015new}. 

\begin{figure*}[t]
\centering
\subfigure[Adaptive Sampling]{
\includegraphics[width=1.4in]{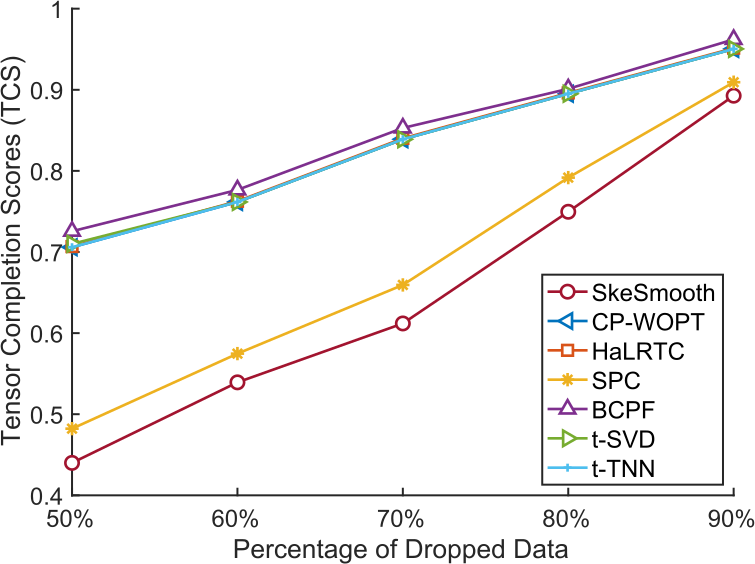}
}
\subfigure[Fixed rate Sampling]{
\includegraphics[width=1.4in]{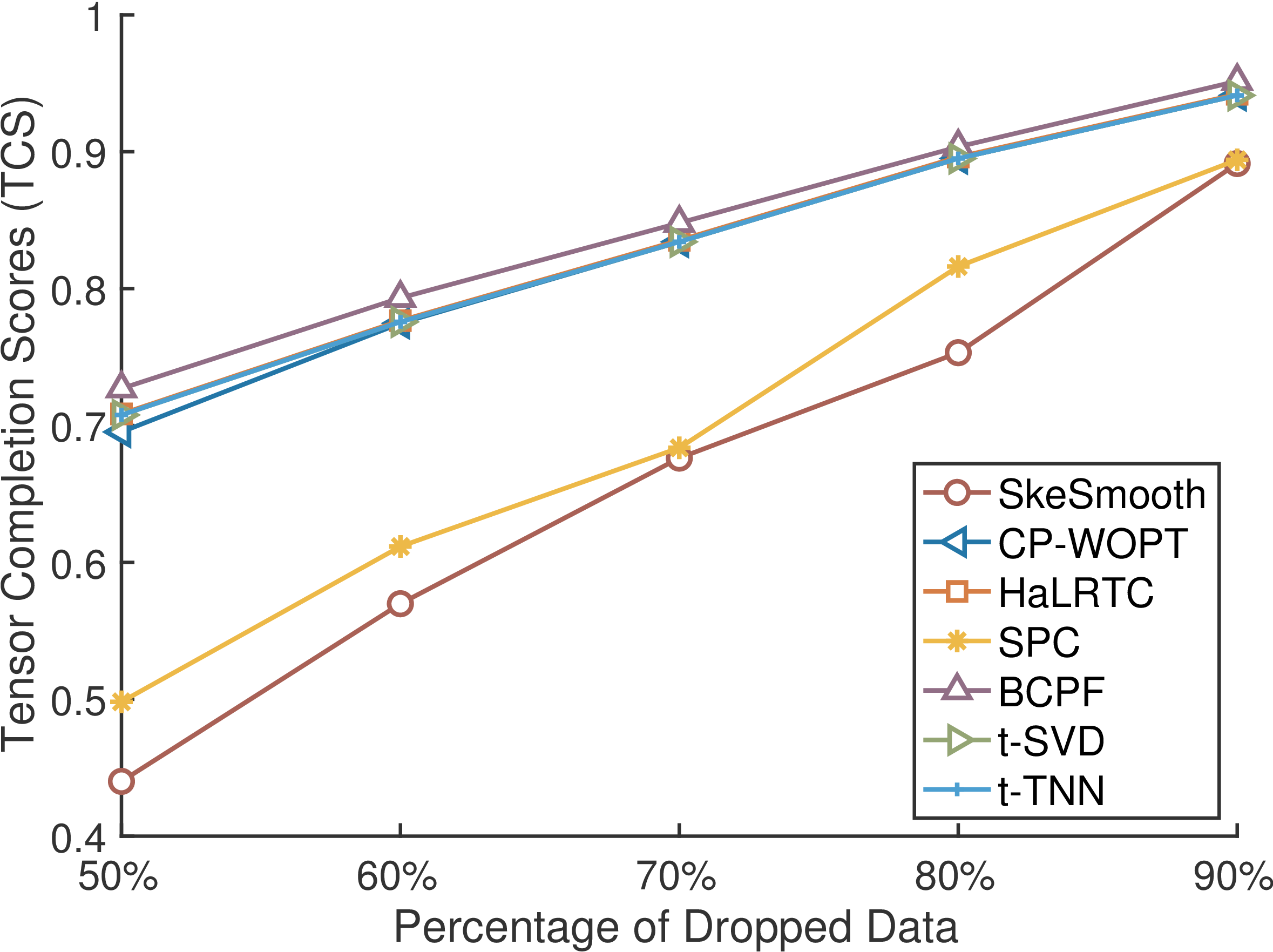}
}
\subfigure[Random Sampling]{
\includegraphics[width=1.4in]{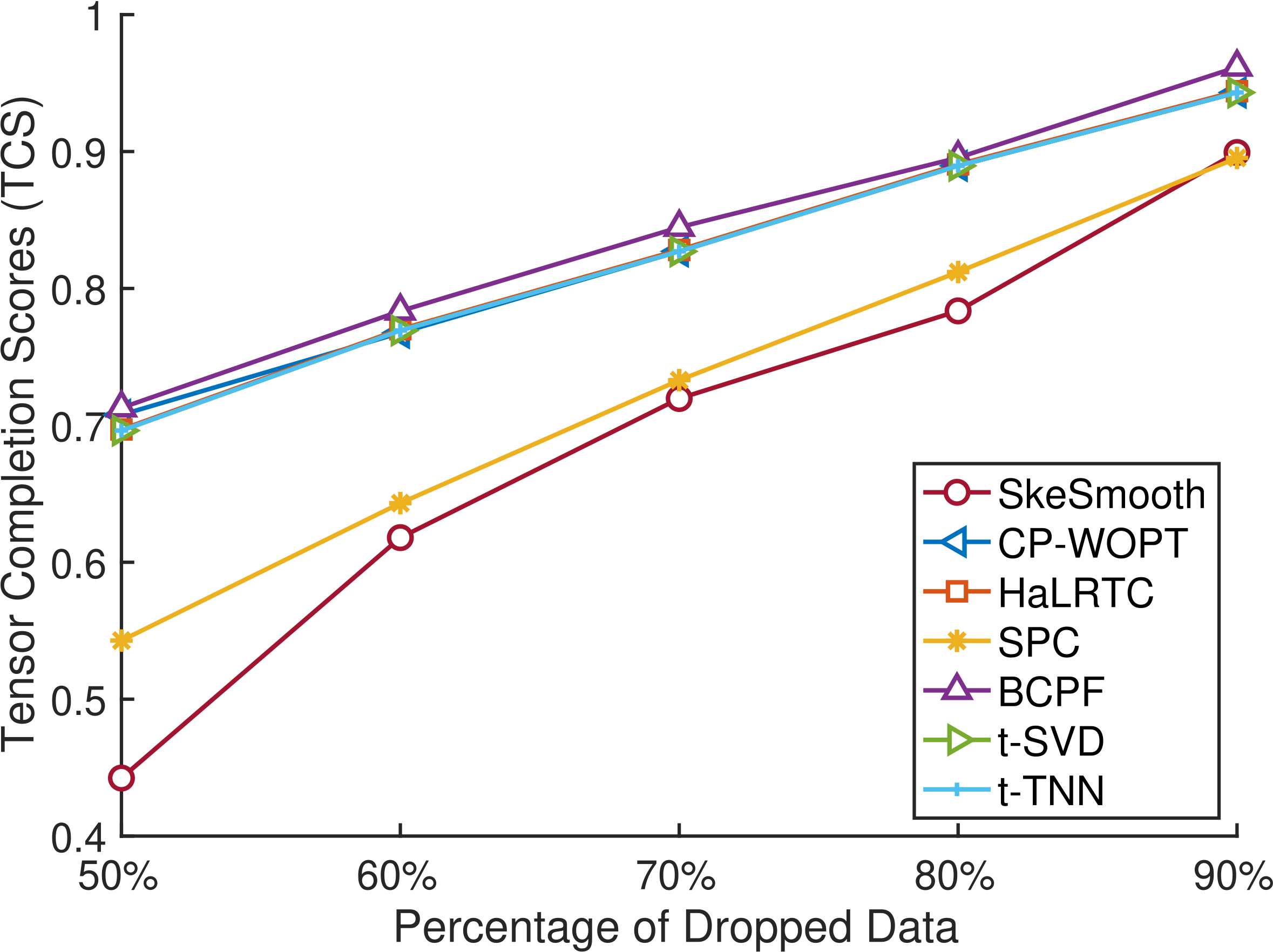}
}
\caption{Tensor Completion Scores Comparison between \methodNameCom~and the Tensor Completion baselines for different sampling mechanisms: (a) adaptive sampling; (b) fixed rate sampling; (c) random sampling.}
\label{fig:tensor_completion}
\end{figure*}

\subsubsection{Evaluation} 
Since not all baseline algorithms are CP-based tensor completions, it is infeasible to compare the FMS. Therefore, we use TCS as the evaluation metric. 
Fig. \ref{fig:tensor_completion} illustrates the TCS as a function of the level of data dropped for different sketching methods. We note that \methodNameCom~outperforms the baseline tensor completion algorithms in achieving a lower TCS (relative error), verifying its robustness in dealing with the missed time slices in the sketched tensor. In particular, we observe the noticeable improvement using the ARIMA-based coefficients in comparison with the SPC algorithm which adopts the QV smoothness constraint. This observation is consistent for all sampling mechanisms.

\subsection{Prediction Task}
It is also critical to analyze how meaningful the generated temporal patterns are by performing a downstream prediction task using the decomposed factor matrices.
We fit a multivariate Long short-term memory (LSTM) model, which is well-suited for the traffic demand prediction task \cite{tsai2009neural}. For the prediction task, the taxi demand is predicted on a daily basis where each day's number of pickups are treated as the daily demand with a train-test partition of 70-30.
The LSTM model uses the factorized temporal mode factor matrix $\textbf{A}$ of size $I\times R$, where each feature within $\textbf{A}$ ($\textbf{A}_{i;}$ of size $1\times R$) represents a latent transportation pattern. 
The LSTM model is configured as one hidden layer of 30 LSTM units with ReLu activation function, followed by an output dense layer\footnote{The model is trained by the RMSprop optimizer with a fixed learning rate (0.01) and a batch size of 8 for 50 epochs.}.

Table \ref{tab:rmse} presents the root-mean-square error (RMSE) differences for adaptive sampling (\methodNameSke), fixed rate sampling, and random sampling with varying percentage of data dropped from 50\% to 90\%. Results demonstrate that adaptive sampling outperforms other two sampling methods in maintaining the predictive power with 60\%-80\% data dropped. As more data are dropped, the test error (RMSE) increases, indicating that it becomes harder to fit the LSTM model. 

\begin{table}[t]
 \centering
\caption{Predictive performance (test RMSEs) for adaptive sampling, fixed rate sampling, and random sampling with 50\% to 90\% of data dropped.}
  \begin{tabular}{ l | p{1.5cm} | p{1.5cm}  | p{1.5cm} | p{1.5cm} | p{1.5cm} }
  \hline
    \textbf{Sampling techniques} & \textbf{50\%} & \textbf{60\%} & \textbf{70\%} & \textbf{80\%} & \textbf{90\%}\\
    \hline
    Adaptive sampling & 0.0155 & \textbf{0.0270} & \textbf{0.0303} & \textbf{0.02827} & 0.0450\\
    Fixed-rate sampling & 0.0173 & 0.0278 & 0.0309 & 0.0427 & 0.0452\\
    Random sampling & \textbf{0.0154} & 0.0282 & 0.0317 & 0.0342 & \textbf{0.0322}\\
    \bottomrule
\end{tabular}
\label{tab:rmse}
\end{table}

\section{Related Works}
We review the previous work regarding the tensor sketching and tensor completion approaches in this section.
\subsection{Tensor Sketching}
Sketching has been proposed and investigated as an indispensable numerical linear algebra tool to help process large volume of data \cite{woodruff2014sketching}. Thus far, there have been two research directions.
Random projection is the predominant form of tensor sketching. Random projection can also be considered as tensor compression via matrix-tensor multiplication.
\cite{sidiropoulos2012multi} developed an approach to randomly compress a big tensor into a much smaller tensor by multiplying the tensor with a random matrix on each mode. \cite{sidiropoulos2014parallel} extend this work to improve the permutation matching performances of the resulting factor matrices. \cite{gujral2018octen} expand the work further to the online setting. However, these compression method cannot avoid the inherent bottleneck of the matrix-tensor multiplication operators. 

An alternative for tensor sketching is tensor sparsification. The goal of tensor sparsification is to generate a sparse ``sketch" of the large tensor. \cite{nguyen2010tensor} proposed random sampling based on the tensor spectral norm. Entries are sampled with probability proportional to the magnitude of the entries. \cite{xia2017effective} extended this work with: 1) the extension from cubic tensors to general tensors; 2) the criteria and treatment of ``small" entry values. They proposed to keep all large entries, sample proportional to moderate entries, and uniformly sample small entries. \cite{bhojanapalli2015new} proposed a new sampling method that samples entries based on a pre-computed tensor distribution. These tensor sparsification algorithms do not consider the data streaming setting and require prior knowledge about tensor. 

\subsection{Tensor Factorization and Completion}
Tensor completion is more considered as a byproduct when dealing with missing data in the tensor factorization process. As a result, CP decomposition and Tucker decomposition are two major types of tensor completion methods \cite{song2019tensor}.  \cite{acar2011scalable} proposed CP-WOPT, a tensor CP decomposition algorithm that can handle missing entries using a binary weight tensor. Tucker-based tensor completion including geomCG \cite{kressner2014low}, HaLRTC/FaLRTC \cite{liu2012tensor}, and TNCP \cite{liu2014trace}, etc. There are other approaches besides CP and Tucker decomposition, such as t-SVD \cite{zhang2014novel}. \cite{song2019tensor} and \cite{sobral_sbmi_prl_2016} provide comprehensive surveys on the topic of tensor completion.

 Tensor completion has also been widely applied to spatio-temporal data analysis for traffic prediction \cite{tan2016short,he2015low}, urban mobility pattern mining \cite{sun2016understanding}, and urban event detection \cite{chen2016fine}. Several works have proposed the tensor completion algorithms for spatio-temporal data. \cite{zhou2015spatio} applied a similar tensor completion algorithm with smoothness constraints as in \cite{yokota2016smooth} and \cite{zhou2015spatio} to the imputation of missing internet traffic data, that tries to minimize the difference between adjacent time slots. \cite{ge2016uncovering} explored the similarity for each element within one dimension and used this as auxiliary information to better recover the original tensor. Nevertheless, the above algorithms did not consider the tensor streaming problem. \cite{tan2016short} proposed a dynamic tensor completion (DTC) algorithm, where traffic data are represented as a dynamic tensor pattern, but still require the entire tensor.

\section{Conclusions}
\frameworkName~is a tensor factorization framework that generates tensor sketches in a streaming fashion using adaptive sampling mechanism and decomposes the sampled smaller tensor sketch with an ARIMA-based smoothness constraint. It is well suited for the incrementally increased spatio-temporal data analysis that can greatly reduce the storage cost and sample complexity, while preserving the latent global structure. \frameworkName~includes two main components: 1) \methodNameSke~as a tensor sketching algorithm that tackles the memory efficiency issue by adaptively adjusting the sampling interval based on the model prediction error; and 2) \methodNameCom~incorporates an ARIMA-based smoothness constraint, which better demonstrates the auto correlation between each time point of a time series. 
Experiments on the large-scale NYC taxi dataset illustrate that with adaptive sampling strategy, \methodNameSke~can greatly reduce the memory cost by 80\% and still preserve the latent temporal patterns and the predictive power. Future work will focus on the distributed setting to improve scalability and efficiency.

\section*{Acknowledgments}
This work was supported by the National Science Foundation,
award IIS-\#1838200. 
\bibliographystyle{splncs04}
\bibliography{bibliography}

\newpage
\section*{Supplementary}
We provide experiment settings details and additional experiment results in the supplementary material.
\subsection{Experiment Settings}
The tensor factorization task is performed using Matlab 2019a on a \texttt{r5.12xlarge} instance of AWS EC2 with Tensor Toolbox Version 2.6 \cite{TTB_Software} for tensor computation and $l-bfgs$ for optimization. For the prediction task, we build the Long short-term memory (LSTM) model using Keras with python 3.6 on a \texttt{p2.xlarge} instance of AWS EC2.

\subsection{Tensor Sketching}
\begin{figure*}[h]
\centering{\includegraphics[width=4.5in]{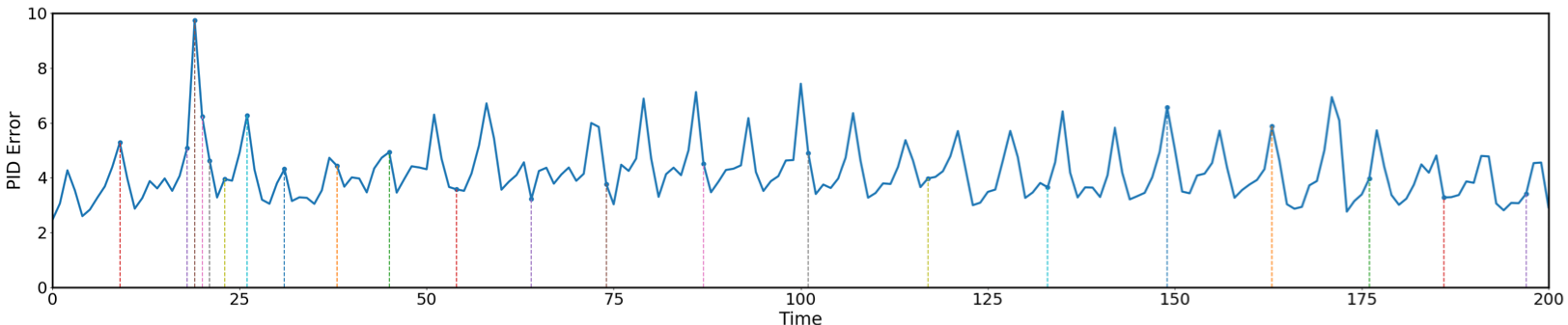}}
\caption{\methodNameSke~sampled timestamps with 90\% data dropped.}
\label{fig:time_sample}
\end{figure*}

Figure \ref{fig:time_sample} illustrates the adaptive sampling results for 90\% data dropped according to the PID errors. To better show the sampling results, we select a portion (200 time slots) from the long streams. From the sampled data points, we observe that for time point at 19-21, the PID error incurs a sudden increase and drop, this reflects on more frequent time slots being sampled at these points. While the error switches to a moderate fluctuation, the sampling interval becomes even.

\begin{figure*}[t]
\centering{\includegraphics[width=4.5in]{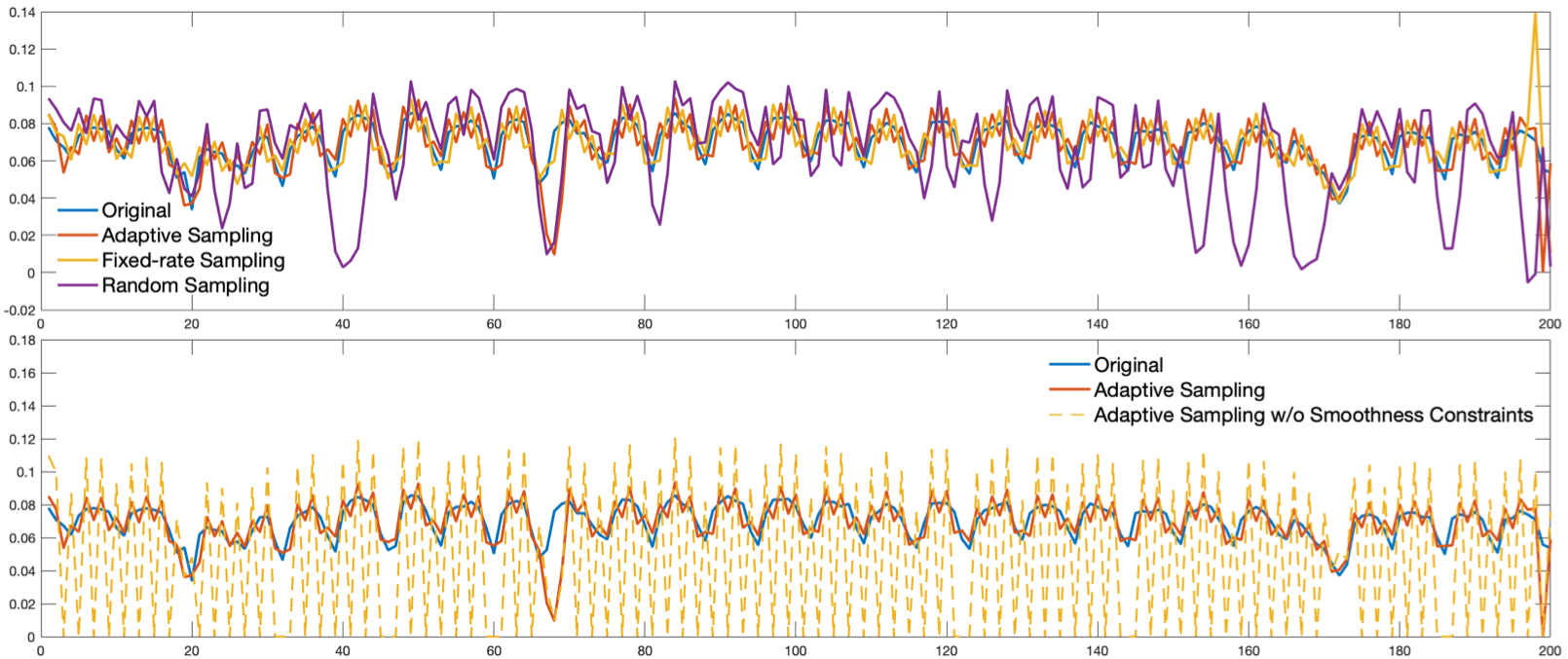}}
\caption{Second component pattern with different level of dropped data for daily based data. The upper figure shows the comparison between different sampling approaches, the lower figure shows the comparison between the full tensor, tensor sketch with \methodNameSke, and the result without smoothness constraints. The x-axis represents the timestamps, y-axis is the value of the components.}
\label{fig:sampling_level}
\end{figure*}

Furthermore, we explore the effect of smoothing in fig. \ref{fig:sampling_level}. We show one of the extracted temporal pattern under the case of 50\% sampled data. Note that we only present 200 time slices to give a better illustration. The upper figure shows the original pattern discovered by CP-OPT \cite{acar2011a} with the complete tensor, compared with the pattern extracted by \methodNameCom~with tensor sketches acquired by different sketching methods. Combining the lower and upper figures, we observe that with the ARIMA-based smoothness constraints, the dropped data points are smoothed out to form a complete and smooth temporal pattern that resembles the original pattern discovered with the full tensors. 
With only 50\% of data remaining, \methodNameCom~is able to recover the original temporal pattern discovered with the full tensor. With more of the time slices dropped, \methodNameCom~will still be able to recover the overall trends, as the values of FMS are still above 0.5 for adaptive and fixed-rate sampling while some of the details might be lost. We notice that adaptive sampling better captures the trend for 50\% data dropped due to the error feedback system, as it samples more frequently at the time points with higher fluctuations.

\begin{figure*}[t]
\centering{\includegraphics[width=4.5in]{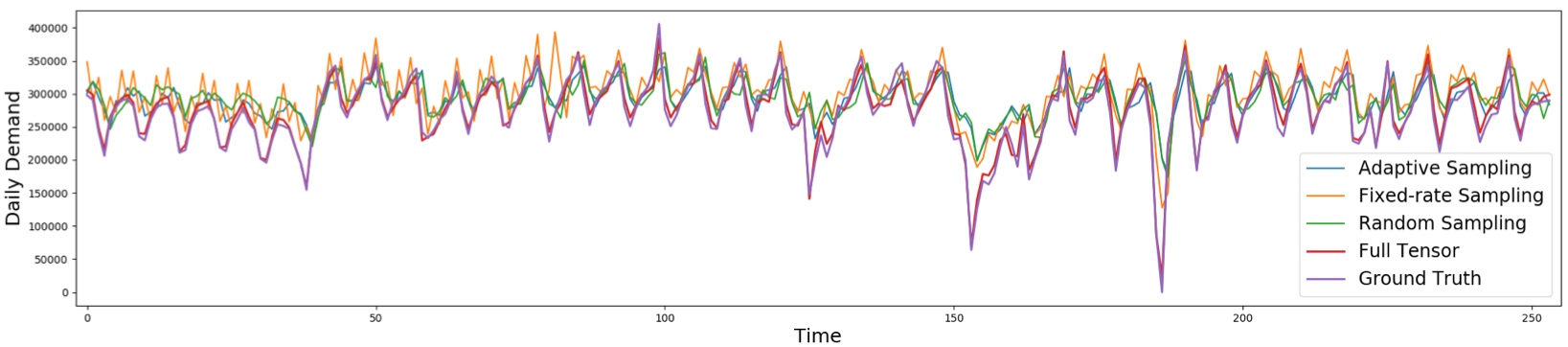}}
\caption{Taxi demand prediction obtained with temporal patterns generated from different sampling methods with 50\% data dropped, compared with the ground truth and the prediction with the temporal pattern extracted from the full tensor.}
\label{fig:lstm_pred}
\end{figure*}

\subsection{Tensor Decomposition with Sketches}
\begin{figure*}[t]
\centering
\subfigure[Adaptive Sampling]{
\includegraphics[width=1.4in]{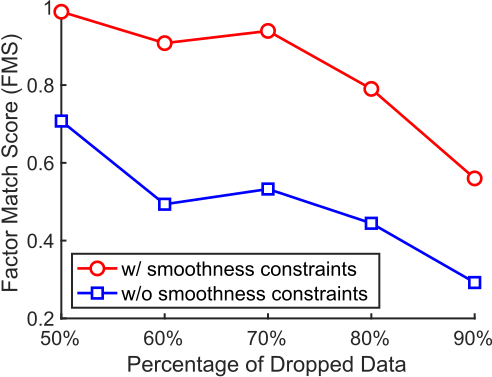}
}
\subfigure[Fixed-rate Sampling]{
\includegraphics[width=1.4in]{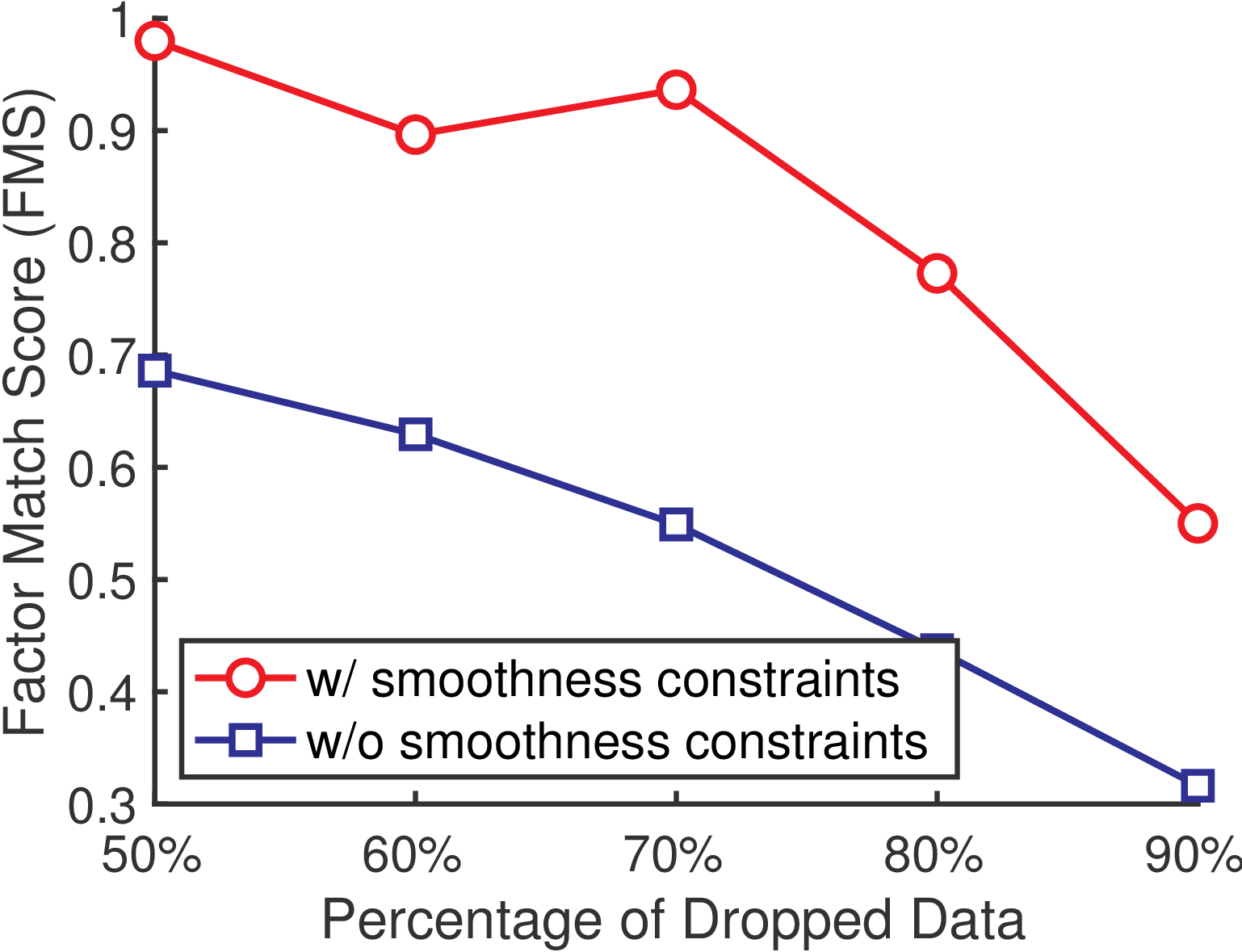}
}
\subfigure[Random Sampling]{
\includegraphics[width=1.4in]{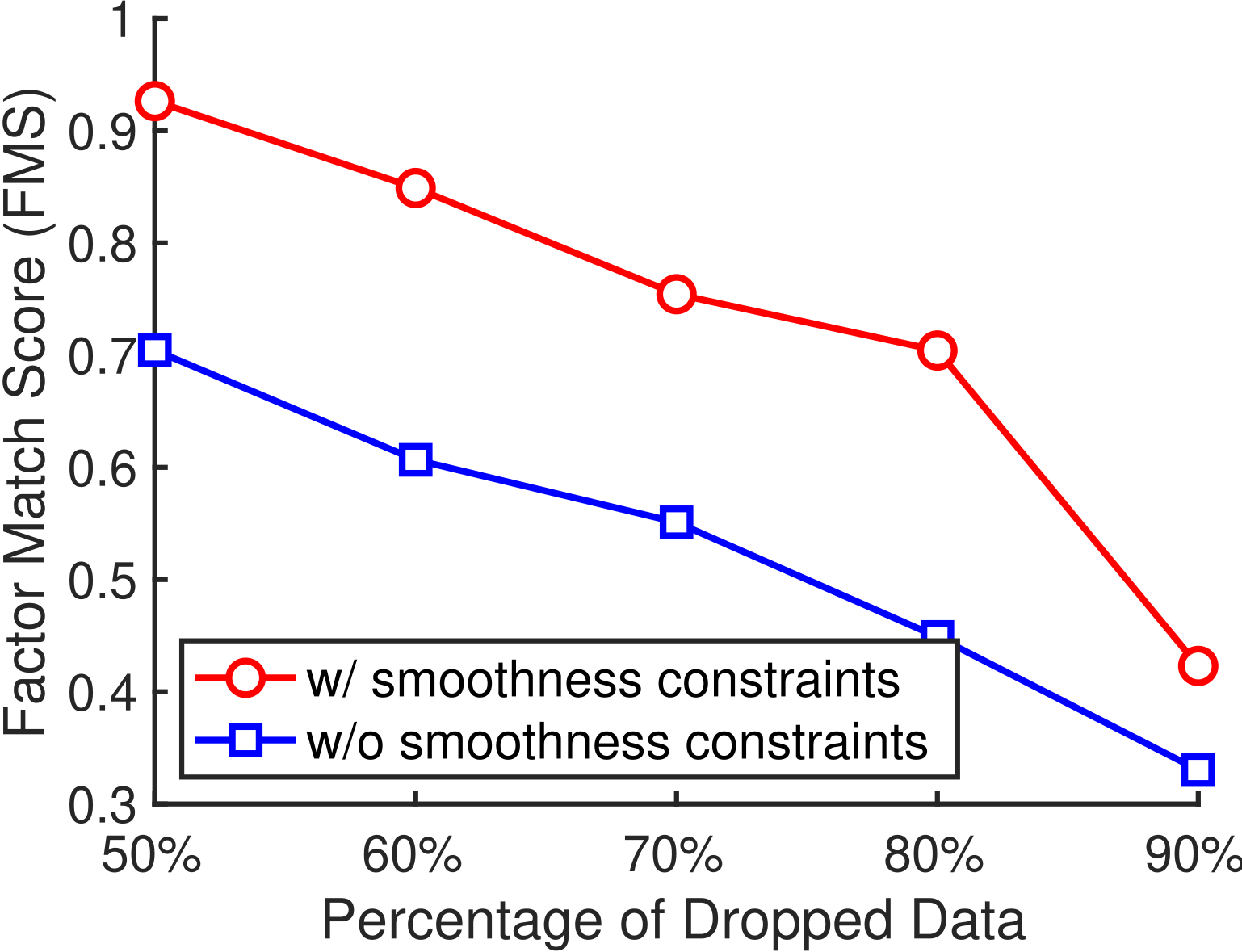}
}
\caption{Factor match score comparison for  \methodNameCom~with and without the smoothness constraints for different sampling mechanisms: (a) adaptive sampling; (b) fixed rate sampling; (c) random sampling.}
\label{fig:fms}
\end{figure*}

\begin{figure*}[htbp]
\centering
\subfigure[Adaptive Sampling]{
\includegraphics[width=1.4in]{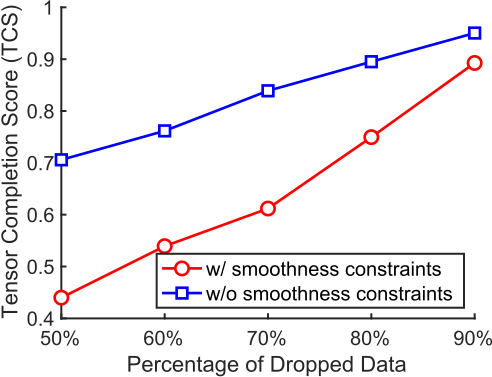}
}
\subfigure[Fixed-rate Sampling]{
\includegraphics[width=1.4in]{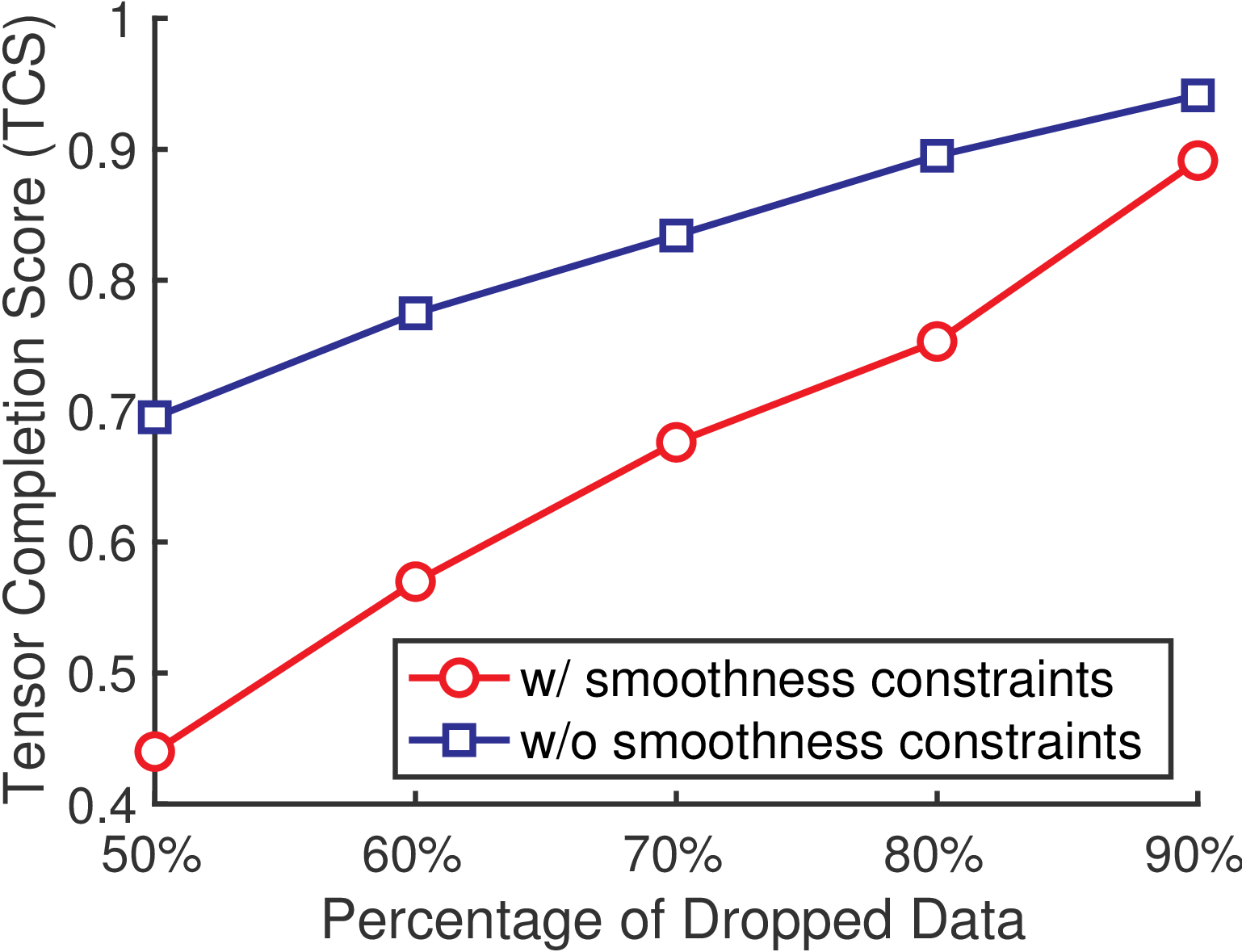}
}
\subfigure[Random Sampling]{
\includegraphics[width=1.4in]{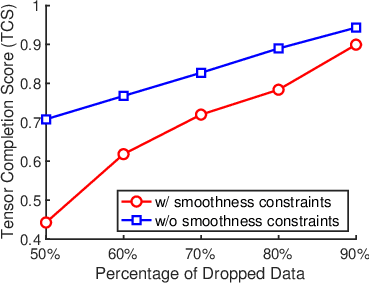}
}
\caption{Tensor completion score comparison for  \methodNameCom~with and without the smoothness constraints for different sampling mechanisms: (a) adaptive sampling; (b) fixed rate sampling; (c) random sampling.}
\label{fig:tcs}
\end{figure*}

Fig. \ref{fig:fms} and \ref{fig:tcs} show the FMS and TCS of different sampling methods with and without the smoothness constraints when applied to \methodNameCom~algorithm. It is obvious that the smoothness constraints achieves significantly better performance as the FMS with smoothness constraints are overall higher than those without any smoothness constraints; and the TCS with smoothness constraints are overall lower than those without the smoothness constraints. 

\subsection{Prediction Task}

Predicting the taxi demand is critical for taxi companies for real-time decision making, monitoring the traffic flow, and helping drivers to determine the most time-saving routes \cite{moreira2013predicting}. 
Fig. \ref{fig:lstm_pred} shows the taxi demand prediction of 254 timestamps with 50\% data dropped by different sampling mechanisms. 
We observe that with only 50\% data from the original data stream, we are still able to extract meaningful temporal patterns that can successfully fit a taxi demand prediction model. 

\subsection{Hourly Basis Anlysis}
\begin{figure*}[htbp]
\centering
\subfigure[50\% Dropped Data]{
\includegraphics[width=2.2in]{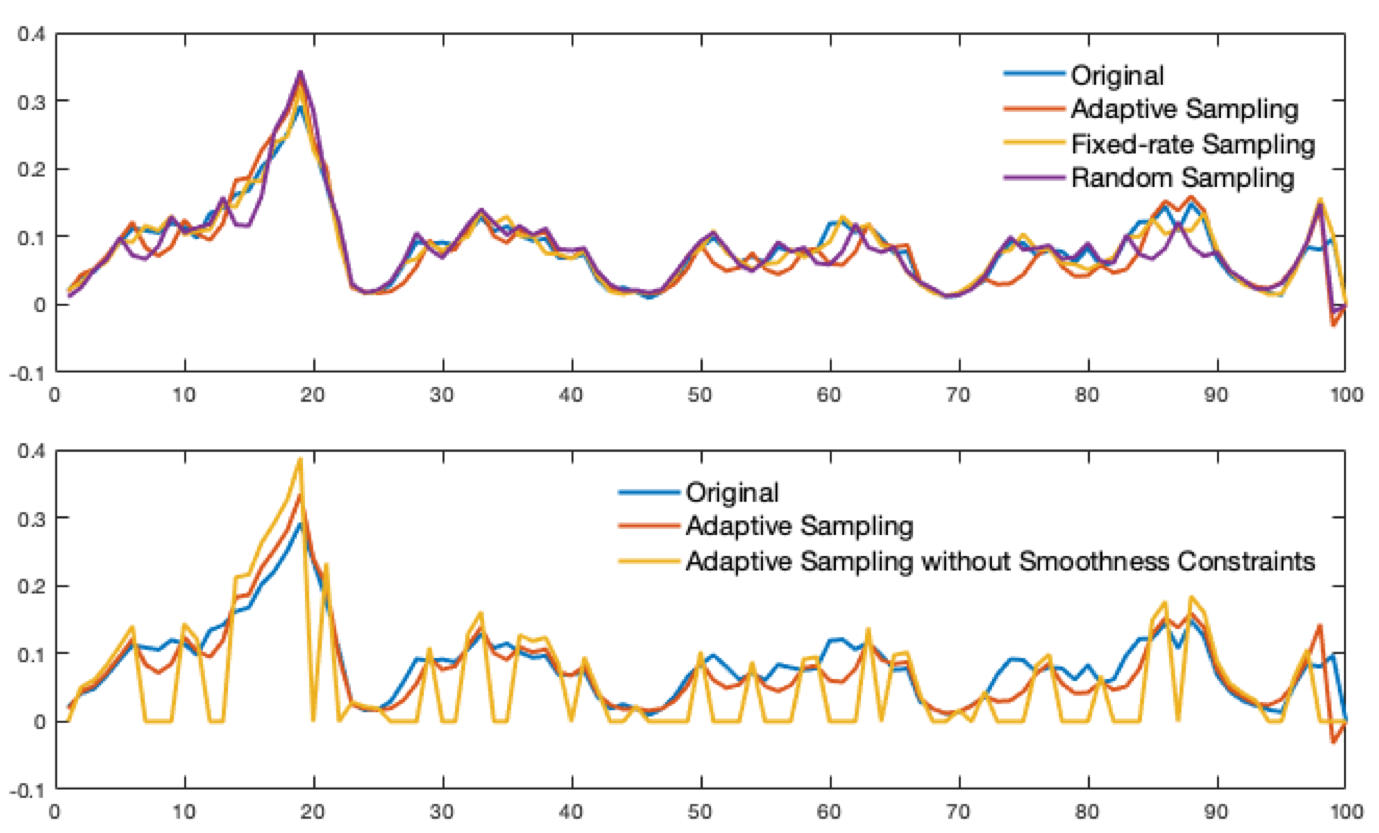}
}
\subfigure[90\% Dropped Data]{
\includegraphics[width=2.2in]{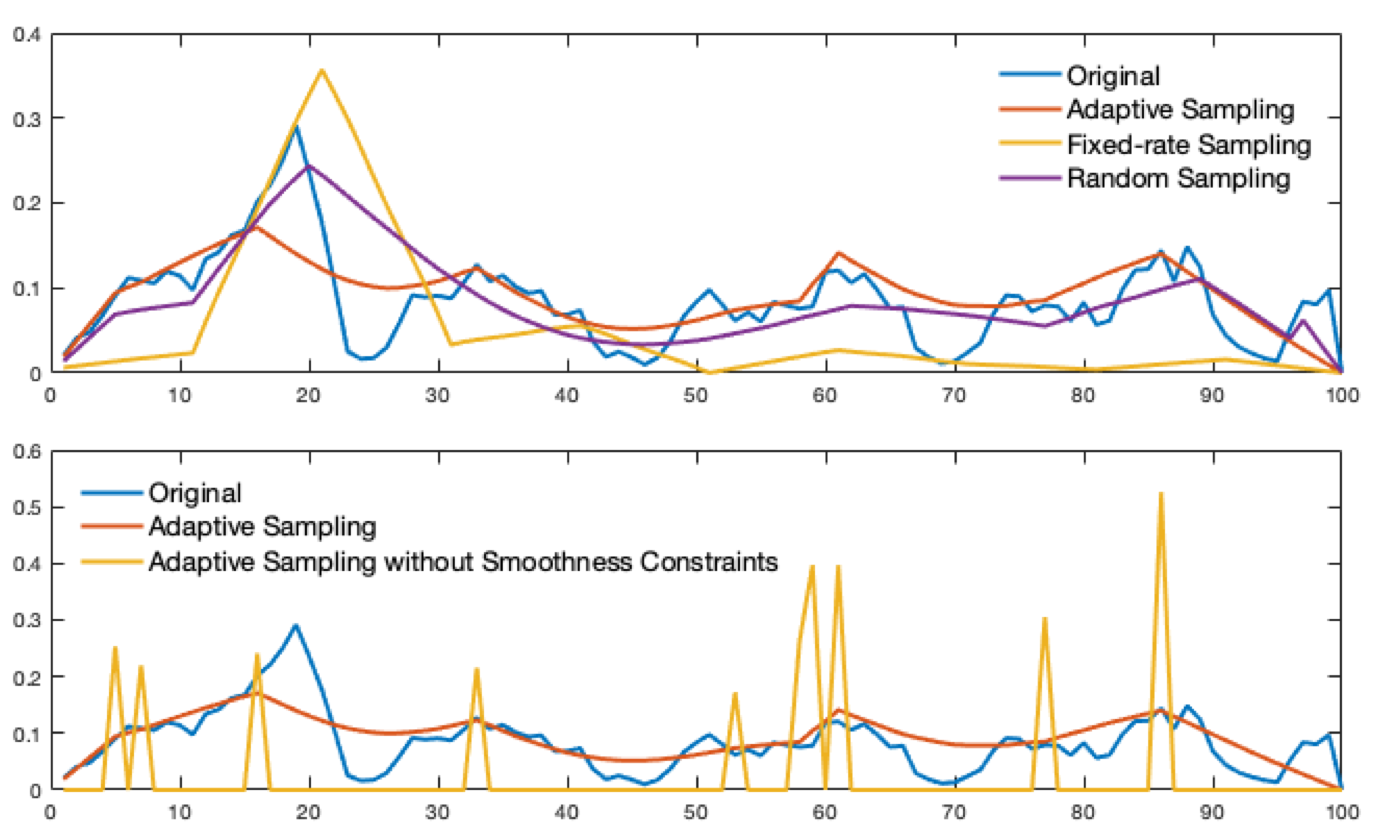}
}
\caption{Second Component pattern with different level of dropped data for hourly based data. The upper figure shows the comparison between the full tensor, tensor sketch with \methodNameSke, and the result without smoothness constraints, the lower figure shows the comparison between different sampling approaches. (a) 50\% dropped data.  (b) 90\% dropped data}
\label{fig:sampling_level2}
\end{figure*}

Besides collecting data on the daily basis, we also show the results based on an hourly collected data. Fig. \ref{fig:sampling_level2} shows the effect of smoothness constraints. We show one of the extracted temporal pattern under two extreme cases: 50\% and 90\% of dropped data. Note that we only present 100 time slices to give a better illustration. It can be seen that consistent with Fig. \ref{fig:sampling_level}, the smoothness constraints play an essential part in recovering smooth patterns. Adaptive sampling (\methodNameSke) achieves the best temporal pattern recovery, while fixed-rate is slightly worse than \methodNameSke. Both adaptive and fixed-rate sampling are better than random sampling method, indicating the necessity of carefully design the sampling strategies.

\begin{figure*}[htbp]
\centering
\subfigure[FMS vs dropped rates]{
\includegraphics[width=1.8in]{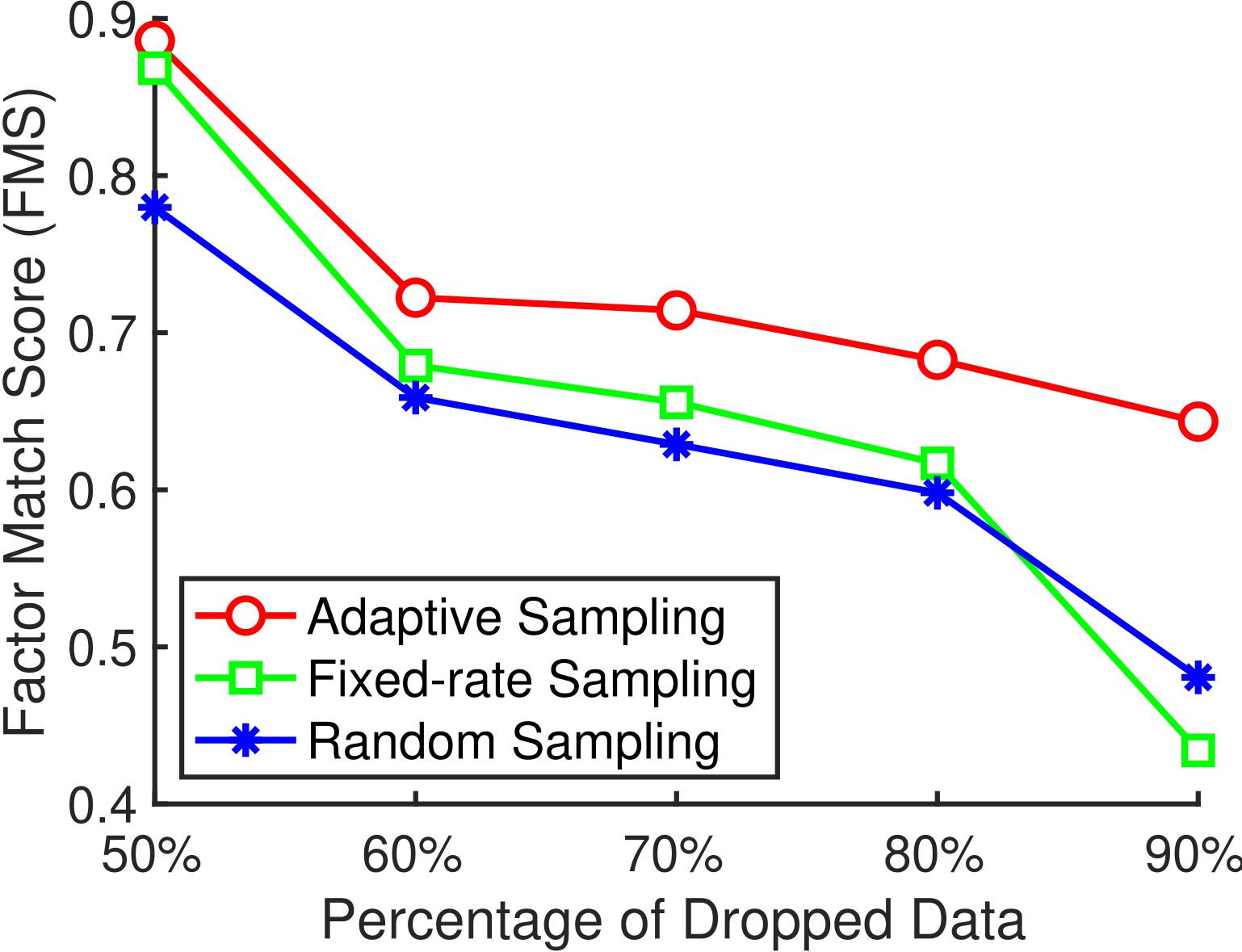}
}
\subfigure[TCS vs dropped rates]{
\includegraphics[width=1.8in]{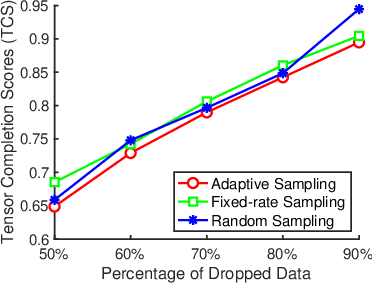}
}
\caption{FMS and TCS over dropped rates ((c) and (d)).}
\label{fig:theta2}
\end{figure*}

\begin{figure*}[htbp]
\centering
\subfigure[Adaptive Sampling]{
\includegraphics[width=1.4in]{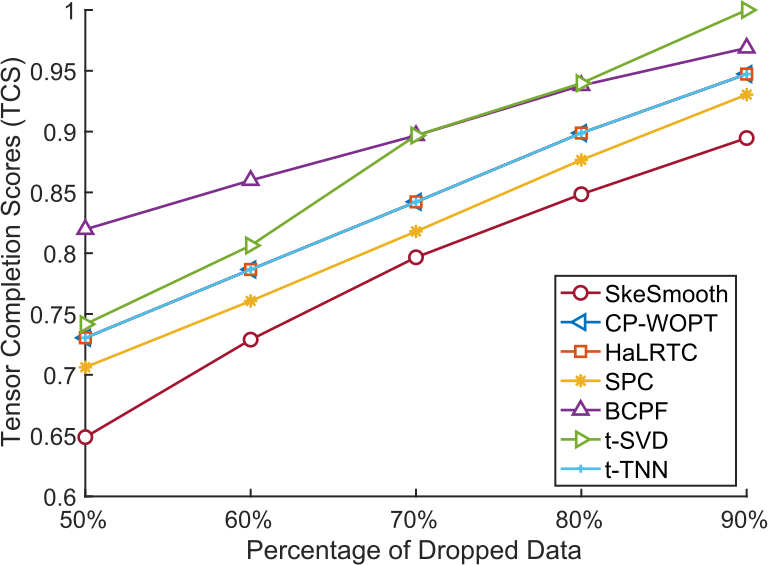}
}
\subfigure[Fixed rate Sampling]{
\includegraphics[width=1.4in]{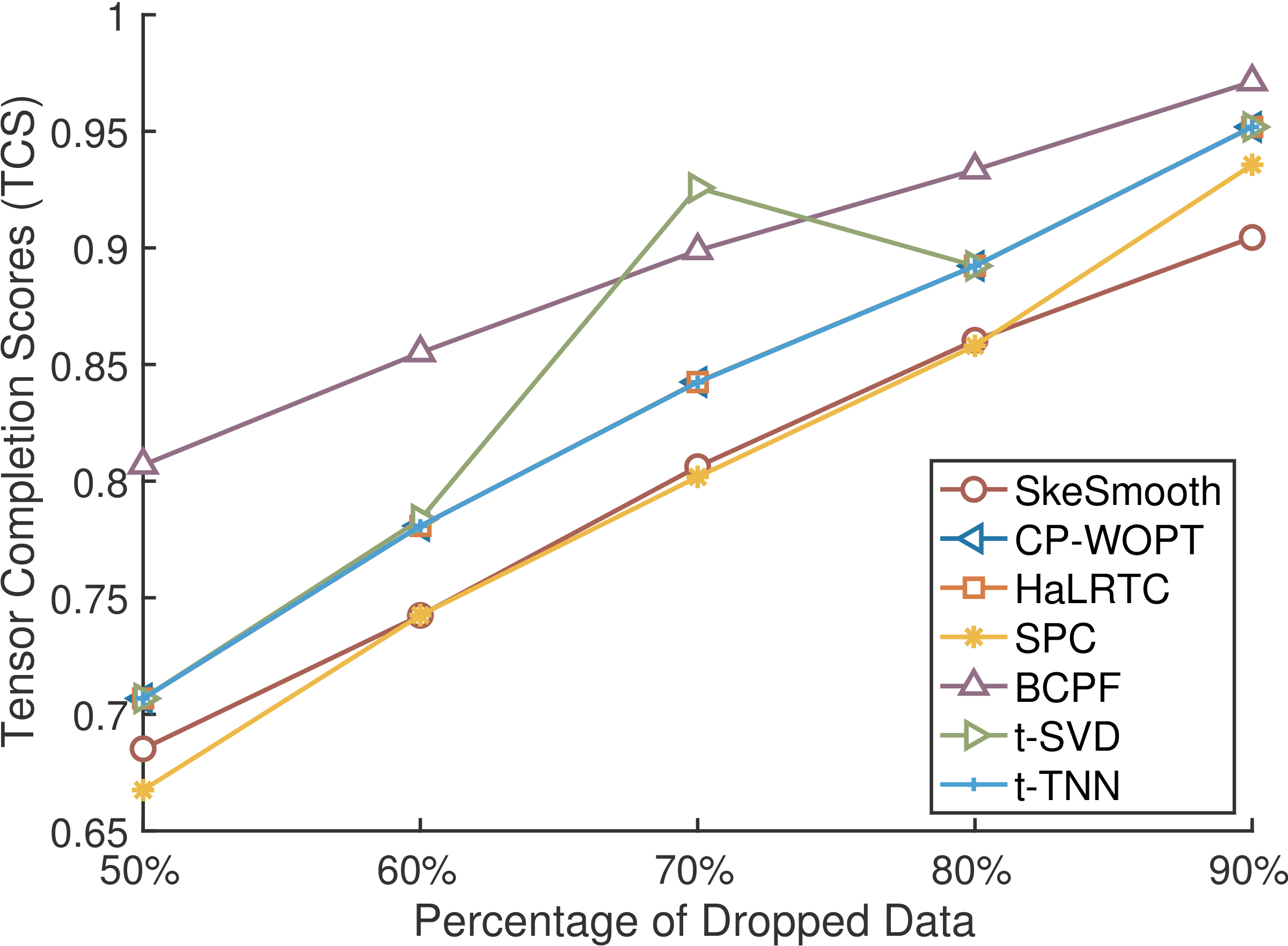}
}
\subfigure[Random Sampling]{
\includegraphics[width=1.4in]{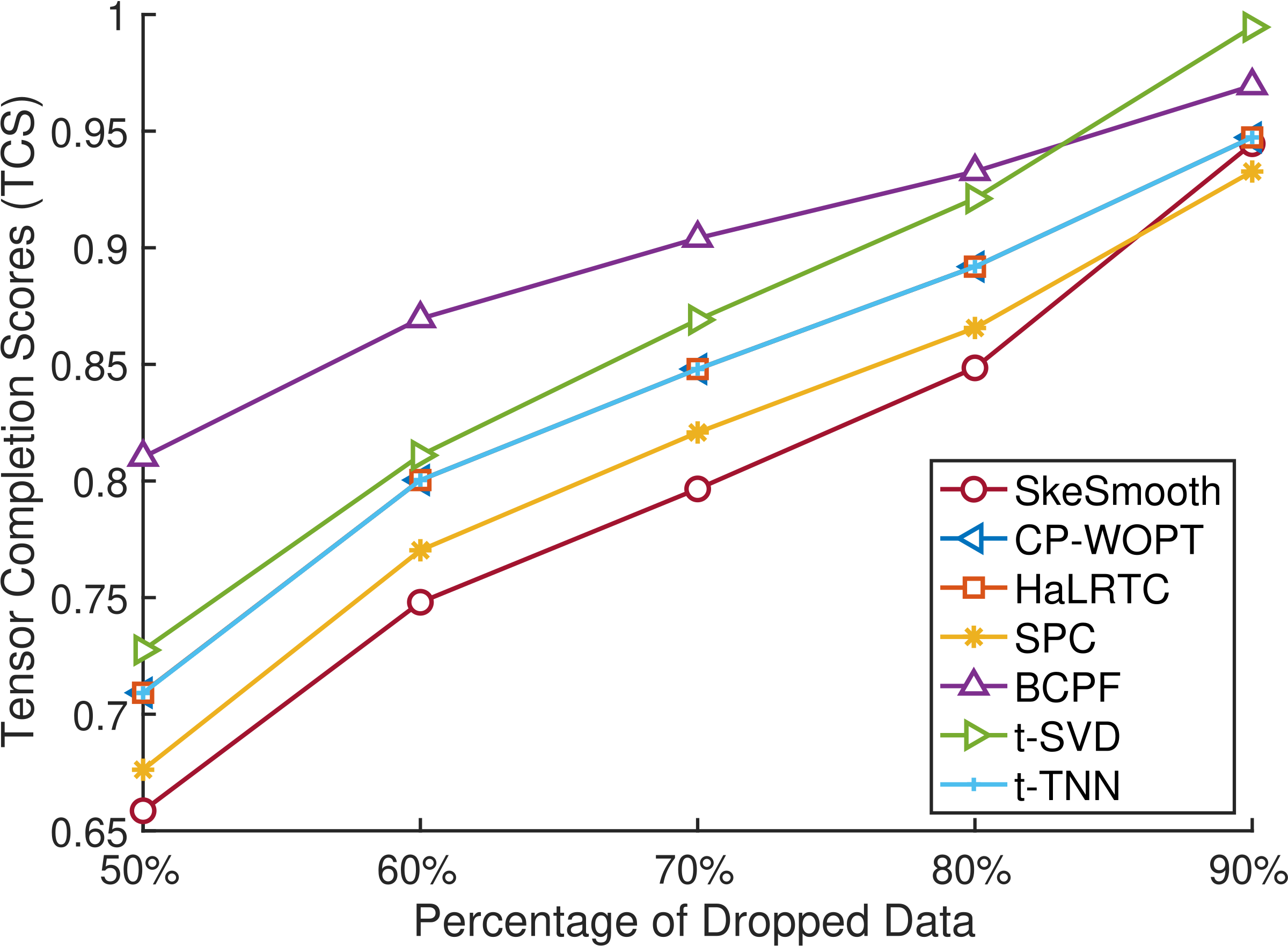}
}
\caption{Tensor Completion Scores Comparison between \methodNameCom~and the Tensor Completion baselines for different sampling mechanisms: (a) adaptive sampling; (b) fixed rate sampling; (c) random sampling.}
\label{fig:tensor_completion2}
\end{figure*}

Fig. \ref{fig:theta2} shows the FMS and TCS with the change of dropped data amount from 50\% to 90\%. Consistent with Fig. 2 (c), (d) in the paper which illustrate the FMS and TCS on data collected on daily basis, they develop the same trend as FMS decrease and TCS increase with more data dropped. Adaptive sampling technique also shows advantageous over the fixed rate sampling and random sampling strategies.
Fig. \ref{fig:tensor_completion2} compares \methodNameCom~with other state-of-the-art baselines. Results demonstrates that \methodNameCom~is more effective and generally has lower recovery errors compared with other baseline methods. The results for both daily basis and hourly basis data collection strategy suggests that our proposed framework \frameworkName~is effective with different kinds of data collection granularity, showing the generalizability of \frameworkName.

\end{document}